%% file: main.tex
\pdfoutput=1

\documentclass[11pt]{article}

\input{_default_preamble}
\input{_custom_preamble}

\title{
Can Language Models Handle a Non-Gregorian Calendar?

The Case of the Japanese \wareki
}

\author{
Mutsumi Sasaki${}^{1}$ \quad
Go Kamoda${}^{2,3}$ \quad
Ryosuke Takahashi${}^{1,4}$ \quad
Kosuke Sato${}^{1}$ \quad
\\
{\bf Kentaro Inui${}^{5,1,4}$ \quad
Keisuke Sakaguchi${}^{1,4}$ \quad
Benjamin Heinzerling${}^{4,1}$ }
\\
${}^1$Tohoku University \quad
${}^2$SOKENDAI \quad
${}^3$NINJAL \quad
${}^4$RIKEN \quad
${}^5$MBZUAI\\
 \small{
   \textbf{Correspondence:} \href{mailto:mutsumi.sasaki@dc.tohoku.ac.jp}{mutsumi.sasaki@dc.tohoku.ac.jp}
 }
}

\begin{document}
\maketitle
\begin{abstract}
Temporal reasoning and knowledge are essential capabilities for language models (LMs).
While much prior work has analyzed and improved temporal reasoning in LMs, most studies have focused solely on the Gregorian calendar.
However, many non-Gregorian systems, such as the Japanese, Hijri, and Hebrew calendars, are in active use and reflect culturally grounded conceptions of time.
If and how well current LMs can accurately handle such non-Gregorian calendars has not been evaluated so far.
Here, we present a systematic evaluation of how well language models handle one such non-Gregorian system: the Japanese \wareki.
We create datasets that require temporal knowledge and reasoning in using \wareki dates. 
Evaluating open and closed LMs, we find that some models can perform calendar conversions, but GPT-4o, Deepseek V3, and even Japanese-centric models struggle with Japanese calendar arithmetic and knowledge involving \wareki dates.
Error analysis suggests corpus frequency of Japanese calendar expressions and a Gregorian bias in the model's knowledge as possible explanations.
Our results show the importance of developing LMs that are better equipped for culture-specific tasks such as calendar understanding.
\footnote{
\includegraphics[width=1em]{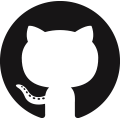}
\href{https://github.com/cl-tohoku/Non-Gregorian-Calendar}{github.com/cl-tohoku/Non-Gregorian-Calendar}
}
\end{abstract}

\section{Introduction}
\begin{figure*}[t!]
    \centering
    \includegraphics[width=\linewidth]{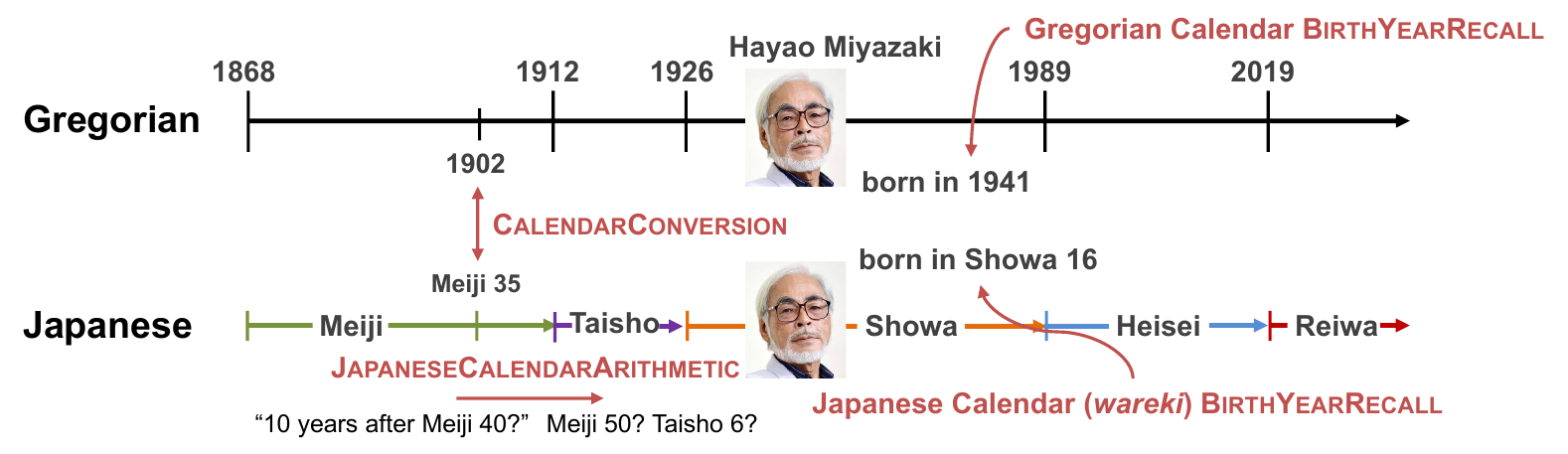}
    \vspace{-4ex}
    \caption{
    In the Japanese calendar (\wareki), years are expressed using era names, which change irregularly according to historic events such as an emperor's accession.
    For example, the Reiwa era began on May 1, 2019, with the accession of Emperor Naruhito, so 2020 corresponds to Reiwa 2.
    In addition to showing the five eras of modern Japan (bottom) in relation to the Gregorian calendar (top), this figure illustrates three tasks designed to evaluate how LMs handle \wareki system:
    (1) \textbf{\textsc{CalendarConversion}} between Gregorian calendar and \wareki;
    (2) \textbf{\textsc{JapaneseCalendarArithmetic}} across era boundaries;
    (3) \textbf{\textsc{BirthYearRecall}} in both calendar systems.
    }
    \label{fig:figure1}
\end{figure*}

The training data of English-centric Language Models (LMs) predominantly assumes the Gregorian calendar as the default temporal framework.
However, many cultures use non-Gregorian calendar systems such as the Islamic \emph{hijri} calendar, the Hebrew calendar, or the Japanese calendar \wareki.
The Hijri calendar guides both religious observances and civil matters in Islamic countries, while the Hebrew calendar remains essential for Jewish religious traditions.
The \wareki system plays an important role in contemporary Japan, appearing in official documents, driver's licenses, and commemorative items. 
Hence, LMs need to be capable of handling such systems to achieve cultural competence.
For example, since the \wareki system is widely used in Japan, LMs encounter \wareki dates in virtually all NLP tasks. For example, a Japanese information retrieval query like ``retrieve all relevant documents from the last 10 years'' might cross era boundaries (\cref{fig:figure1})
and in cross-cultural settings involving different calendar systems, tasks like machine translation and cross-lingual information retrieval require translating dates across calendars.

Although the importance of incorporating cultural perspectives into language models is increasingly recognized~\citep{shen-etal-2024-understanding, pawar2024surveyculturalawarenesslanguage} and efforts have been made to build cultural commonsense benchmarks across languages and regions~\citep{khairallah-etal-2024-camel, kim-etal-2024-click, wang-etal-2024-kulture}, little attention has been paid to culturally grounded temporal expressions such as calendars.
Recent work has evaluated date arithmetic~\citep{wang-zhao-2024-tram, gaere2025datetimenewbenchmarkmeasure, chu-etal-2024-timebench}, temporal reasoning~\citep{NEURIPS_DATASETS_AND_BENCHMARKS2021_1f0e3dad}, date format understanding~\citep{bhatia2025datefragmentshiddenbottleneck, bhatia-etal-2025-datelogicqa}, and has analyzed the impact of tokenization~\citep{bhatia2025datefragmentshiddenbottleneck} and internal representations~\citep{heinzerling-inui-2024-monotonic, el-shangiti-etal-2025-geometry} on calendar-based reasoning in LMs, but these efforts exclusively used the Gregorian calendar.

Here, we focus on \wareki as a representative non-Gregorian system, motivated by its widespread use in Japan, its relative complexity, and the availability of both English- and Japanese-centric open models for cross-linguistic comparison.
The \wareki system divides time into eras, each starting from year 1. Since the end of an era is tied to major historical events such as imperial succession, their lengths are irregular; Meiji, Taisho, Showa, Heisei lasted 45, 15, 64, and 31 years, respectively, and the end of the current Reiwa is unknown. 
A further complication is that a Gregorian year can span two eras if an era transition occurs in that year.
For example, the year 2019 corresponds to Heisei 31 until April 30 and then becomes Reiwa 1 from May 1. 
In non-transition years, a Gregorian year maps exactly to one \wareki year, such as 2020 = Reiwa 2.

Given these properties of \wareki, we designed three evaluation tasks in English and Japanese, focusing on both factual knowledge and reasoning (\cref{fig:figure1}).
Our evaluation of four English-centric open-source models, five Japanese-centric open-source models, and two frontier models shows that English-centric models face considerable difficulties with conversions and reasoning.
In contrast, Japanese-centric models and the frontier models demonstrated relatively good performance in a simple format conversion task. Similarly, they struggled with more complex tasks, such as reasoning across era transitions and recalling birth years in the Japanese calendar.
Furthermore, our error analysis revealed that the accuracy of the \wareki reasoning task for each era is strongly correlated with the corpus frequency of \wareki expressions, and that the performance in the \wareki recall task is influenced by the Gregorian bias of the model's knowledge.
Considering the widespread use of \wareki by over 100 million people in Japan and Japanese being a high-resource language, our findings highlight a surprisingly low capability in Japanese-centric LMs.
In a broader context, we hope to encourage further work aimed at evaluating and improving culture-specific temporal reasoning.

\section{How well do English-centric and Japanese-centric LMs handle \wareki?}

To evaluate LMs' ability to handle the Japanese calendar \wareki, we design three tasks that target distinct aspects of calendar reasoning: \textbf{\textsc{CalendarConversion}} (\cref{sec:calendar_conversion}), \textbf{\textsc{JapaneseCalendarArithmetic}}(\cref{sec: Japanese_calendar_arithmetic}), and \textbf{\textsc{BirthYearRecall}}(\cref{sec:birth_year_recall}).
Our analysis, based on the synthetic data we created (see \cref{sec:datasets} for details), covers the five Japanese eras from 1868 to the present: Meiji, Taisho, Showa, Heisei, and Reiwa.
We evaluate eleven language models in total, including five Japanese-centric models (llm-jp-3-13b~\citep{llmjp2024llmjpcrossorganizationalprojectresearch}, sarashina2-13b, Swallow-13b~\citep{swallow_01, swallow_02}, Swallow-MS-7b, and Llama3-Swallow-8B), four English-centric models (Llama-2-7B, Llama-2-13B~\citep{touvron2023llama2openfoundation}, Mistral-7B~\citep{jiang2023mistral7b}, and Llama3-8B~\citep{grattafiori2024Llama3herdmodels}), and two frontier models, GPT-4o~\citep{openai2024gpt4ocard} and DeepSeek V3~\citep{deepseekai2025deepseekv3technicalreport}(models details in \cref{sec:models}).
Japanese-centric LMs are prompted in Japanese, while English-centric LMs are prompted in English, using few-shot prompts to encourage format adherence .
For GPT-4o and DeepSeek V3, we use both Japanese and English few-shot prompts
(prompt details in \cref{sec:prompts}).

We adopted greedy decoding with deterministic outputs.
For \textsc{CalendarConversion}, we used a single prompt (since paraphrases did not appear to have any impact in preliminary experiments), while for \textsc{\textsc{JapaneseCalendarArithmetic}}, and \textsc{BirthYearRecall}, we used multiple prompt variants and report aggregated scores.

\begin{figure}[t!]
\includegraphics[width=\linewidth]{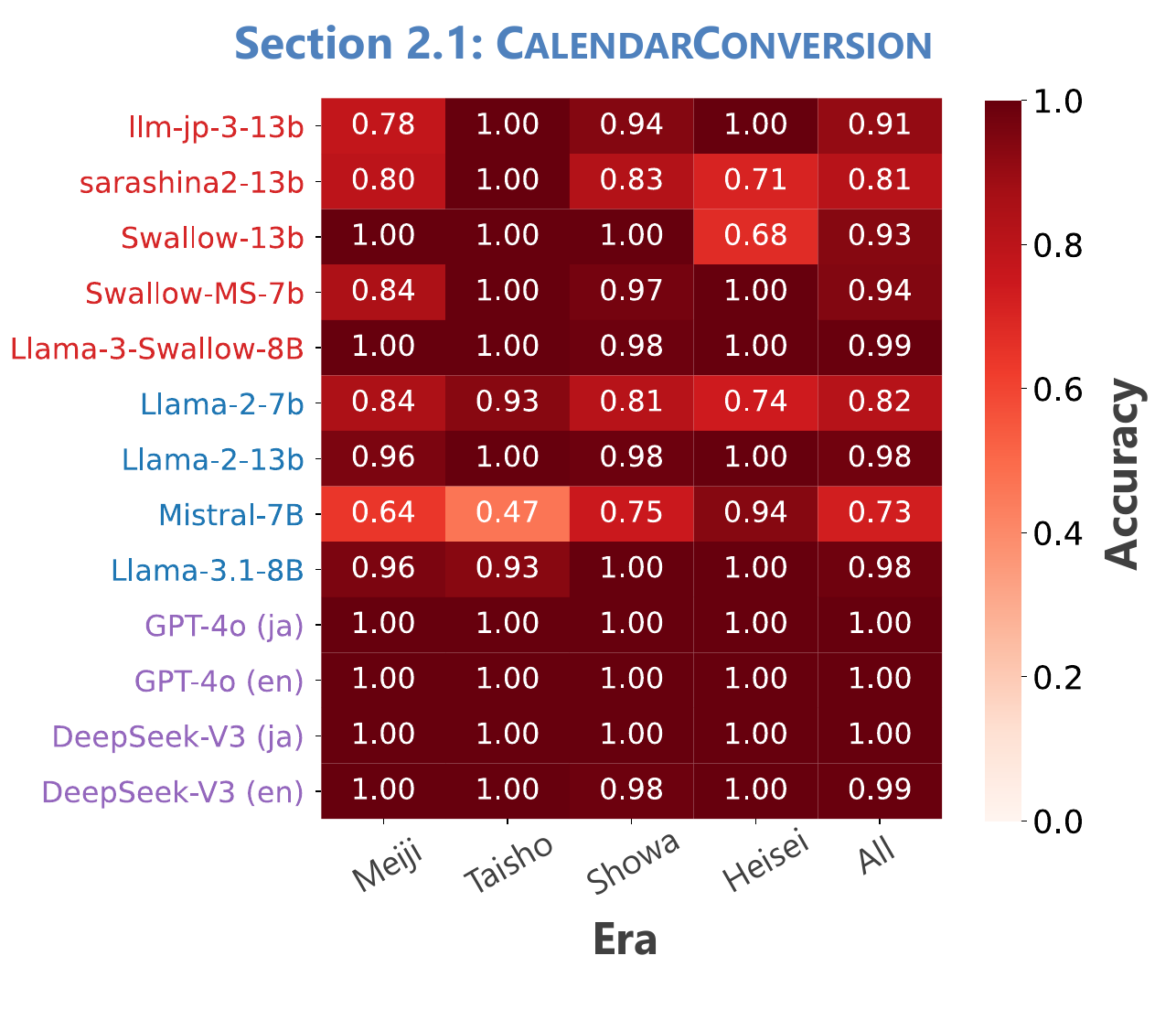}
\caption{
Performance on \textsc{CalendarConversion} (Gregorian→Japanese setting).
Japanese-centric LMs (\textcolor{heatmapred}{red labels}) and frontier LMs (\textcolor{heatmappurple}{purple labels}) perform nearly perfectly across all eras.
Some English-centric LMs (\textcolor{heatmapblue}{blue labels}) fail even at simple conversions.
}
\label{fig:conversion}
\end{figure}

\subsection{\textsc{CalendarConversion}}\label{sec:calendar_conversion}
\paragraph{Settings.}
This task evaluates the ability to convert years between the Gregorian calendar and \wareki.
We constructed a dataset of corresponding Gregorian and \wareki years for the Meiji (1868–1912), Taisho (1912–1926), Showa (1926–1989), and Heisei (1989–2019) eras.
In this task, an LM prompted ``In the Japanese calendar, 1804 corresponds to Bunka 1. In the Japanese calendar, 1992 corresponds to'' should output ``Heisei 4''.\footnote{For fairness across eras, one-shot examples are from eras not included in our evaluation, e.g., Bunka, Koka, or Tenpo.}

For each era, we measure conversion accuracy in both directions.
For Gregorian targets, accuracy is defined as: 
$\frac{1}{N} \sum_{i=1}^N \mathbbm{1}(\hat{y}_i = y_i)$, 
where $\hat{y}_i$ and $y_i$ are the predicted and target Gregorian year for instance $i$, $N$ the number of instances, and $\mathbbm{1}$ the indicator function.
For \wareki{} targets, accuracy is defined as:
$\frac{1}{N} \sum_{i=1}^N \mathbbm{1}(\hat{E}_i = E_i \land \hat{x}_i = x_i)$,
where $E_i$ and $x_i$ are the target era and year in the era (e.g. ``Heisei'' and ``4'' in ``Heisei 4''), and $\hat{E}_i$ and $\hat{x}_i$ are the corresponding model predictions.

\paragraph{Results.}
\cref{fig:conversion} shows the accuracy of Gregorian-to-Japanese \textsc{CalendarConversion}.
Japanese-centric models, GPT-4o, and DeepSeek V3 consistently achieved near-perfect accuracy across all eras, demonstrating strong conversion capability.
In contrast, English-centric LMs show large variations in performance across models and eras. For example, 13B English LMs achieved over 90\% accuracy across all eras, whereas Llama-2-7b and Mistral-7B achieved much lower accuracy.

\begin{figure}[t!]
\includegraphics[width=\linewidth]{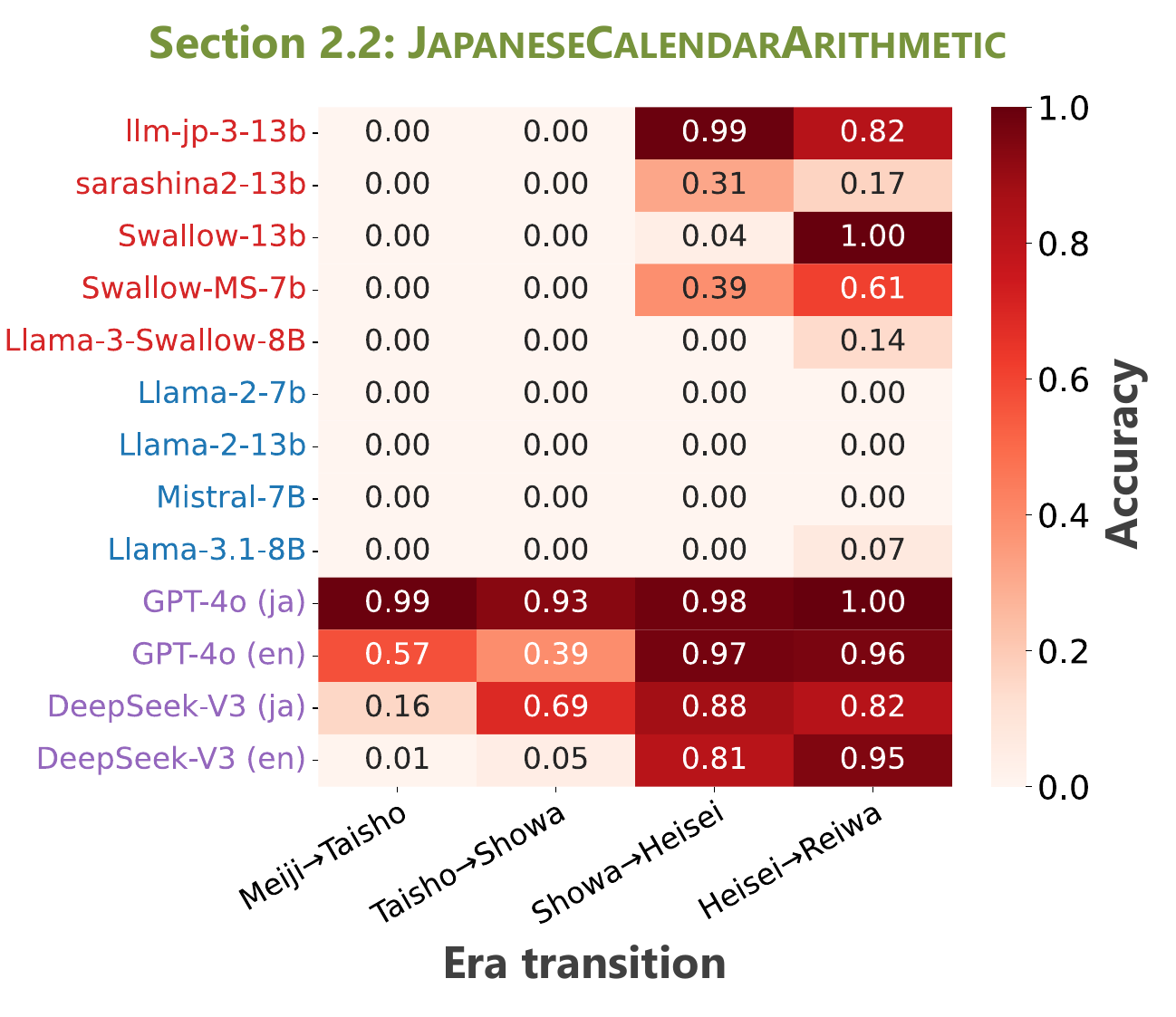}
\caption{
Performance on \textsc{\textsc{JapaneseCalendarArithmetic}}.
A large performance gap is observed between Japanese-centric LMs (\textcolor{heatmapred}{red labels})  and English-centric LMs (\textcolor{heatmapblue}{blue labels}).
Even frontier LMs (\textcolor{heatmappurple}{purple labels}) struggle with this task.
}
\label{fig:arithmetic}
\end{figure}

\subsection{\textsc{\textsc{JapaneseCalendarArithmetic}}}\label{sec: Japanese_calendar_arithmetic}
\paragraph{Settings.}
This task evaluates the ability to perform date arithmetic across \wareki era boundaries.
Specifically, we select a date within a five-year window before each era transition, as well as the date ten years after.
We sampled 500 unique and non-overlapping dates for each era.
In this task, LMs are required to answer the year that is ten years after the given input date.
Each instance is constructed so that an era transition always occurs, from the input-side era (referred to as the pre-era) to the output-side era (referred to as the post-era).
For example, when prompted ``Ten years after March 8, Tenpo 14 is March 8, Koka 3. Ten years after September 19, Heisei 27 is'' the LM should answer ``September 19, Reiwa 7''. 
We evaluate models using the accuracy. 
It is defined as the ratio of outputs that exactly match the correct date:
$\frac{1}{N} \sum_{i=1}^N \mathbbm{1}(E_i = \hat{E}_i \land x_i = \hat{x}_i)$.

\begin{figure}[t!]
\includegraphics[width=\linewidth]{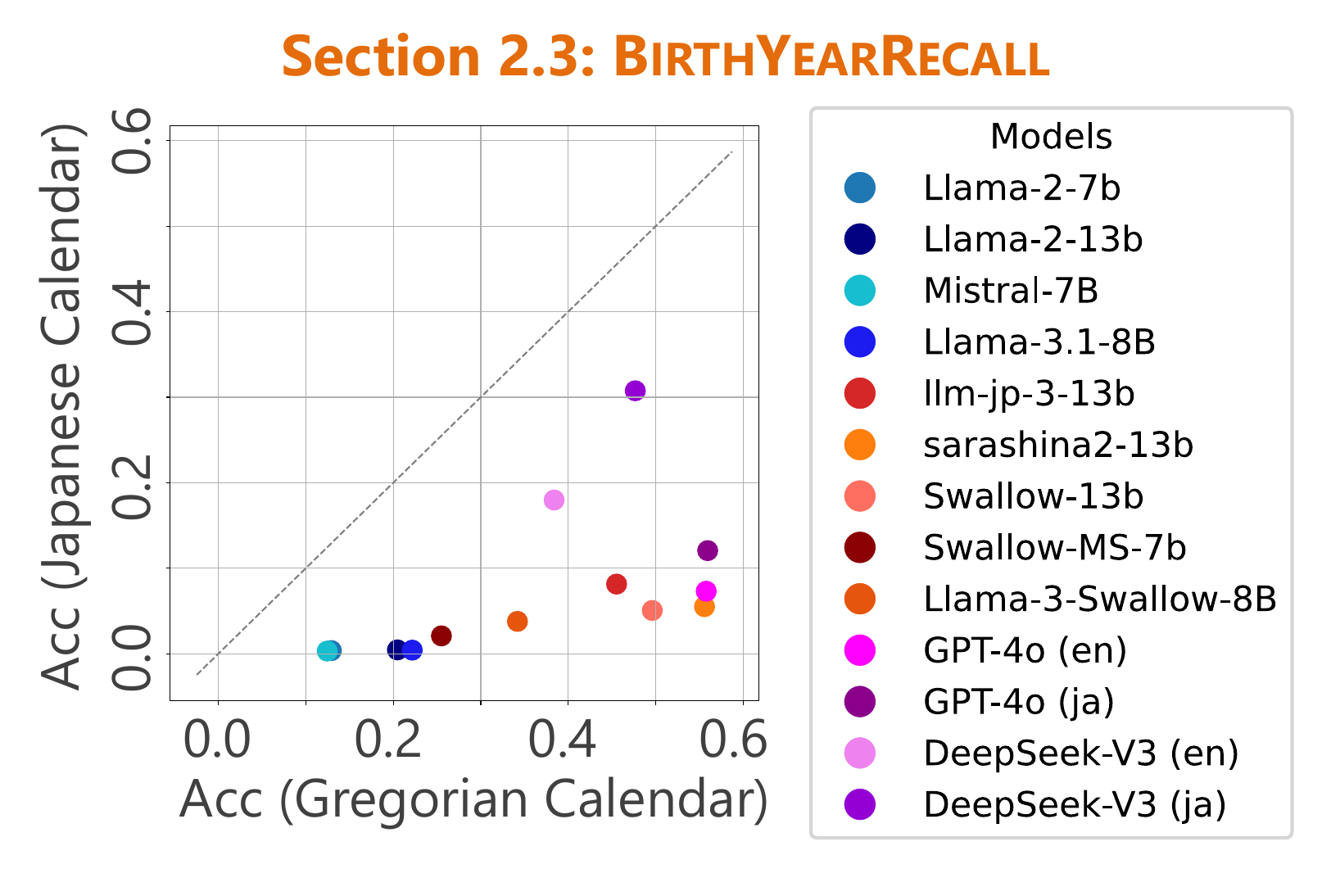}
\centering
\caption{
Comparison of \textsc{BirthYearRecall} accuracy in both Gregorian and \wareki formats.
The diagonal marks equal accuracy; below it indicates a Gregorian bias.
Even Japanese-centric LMs and frontier LMs exhibit a strong bias towards the Gregorian calendar, and Japanese-centric LMs perform comparatively better with the Japanese calendar than English-centric LMs.
}
\label{fig:human_recall}
\end{figure}

\paragraph{Results.}
\cref{fig:arithmetic} shows the results for \textsc{\textsc{JapaneseCalendarArithmetic}}. 
All Japanese-centric and English-centric LMs showed low accuracy on early era transitions (Meiji→Taisho and Taisho→Showa).
Even a frontier LM, DeepSeek V3,  struggled with the era transitions of these era pairs.
In contrast, for more recent transitions such as Heisei→Reiwa and Showa→Heisei, Japanese-centric LMs demonstrated notably better performance.
For example, llm-jp-3-13b achieved accuracies of 0.99 and 0.82, respectively.
On the other hand, English-centric LMs showed almost zero or very low accuracy even for recent eras.

Evaluation using a more lenient metric (\cref{sec:appendix_arithmetic}) revealed that the performance gap between Japanese and English models mainly stems from their handling of same-year transitions (e.g., Heisei 31 → Reiwa 1 in 2019).
Specifically, English models often fail to recognize such transitions, whereas Japanese models tend to handle them correctly.

\subsection{\textsc{BirthYearRecall}}\label{sec:birth_year_recall}
\paragraph{Settings.}
This task measures the ability to recall the birth year of Japanese individuals.
We extracted 300 Japanese individuals per era from Wikidata, filtering for entities with at least 20 relations. 
For example, given the three-shot prompt ``According to the Japanese calendar, Ieyasu Tokugawa was born in Tenmon 11.
($\cdots$).
According to the Japanese calendar, Mao Asada was born in'' the model should answer ``Heisei 2''.
We evaluate models using accuracy, which is an exact match of the prediction and the target.
For Gregorian output, accuracy is
$\frac{1}{N} \sum_{i=1}^N \mathbbm{1}(\hat{y}_i = y_i)$.
For \wareki output, the prediction must match both the era and the year within the era:
$\frac{1}{N} \sum_{i=1}^N \mathbbm{1}(E_i = \hat{E}_i \land x_i = \hat{x}_i)$.

\begin{figure}[t!]
\centering
\includegraphics[width=\linewidth]{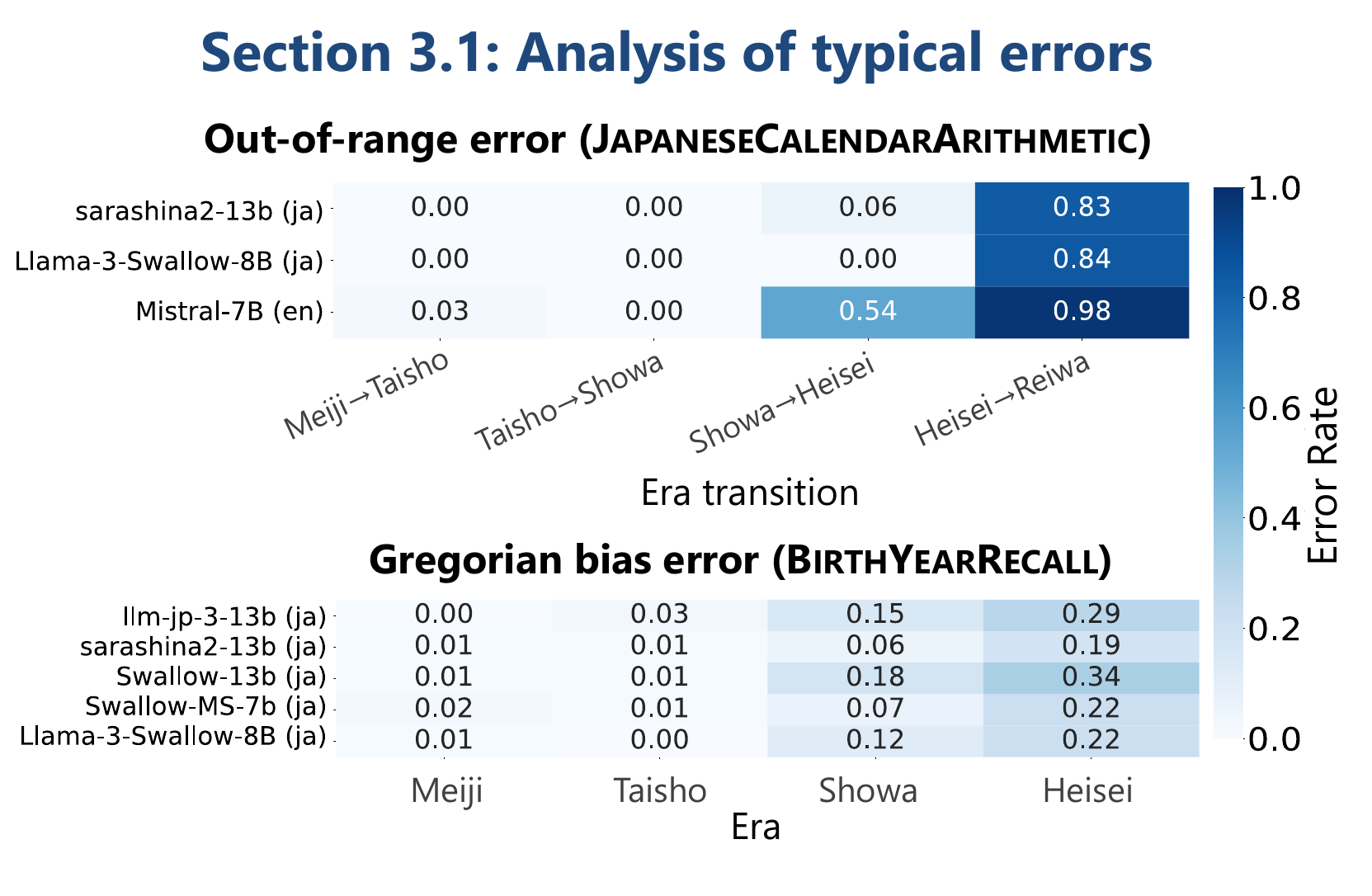}
\caption{
Analysis of why many models fail from the perspectives of typical error patterns.
In \textsc{\textsc{JapaneseCalendarArithmetic}}, out-of-range errors (e.g., generating ``Heisei 37'') may contribute to failures in newer eras.
In \textsc{BirthYearRecall}, Gregorian bias errors (responding in Gregorian years despite 3-shot \wareki prompts) cause failures, especially in newer eras.
}
\label{fig:typical_error}
\end{figure}

\begin{figure*}[t!]
\includegraphics[width=\linewidth]
{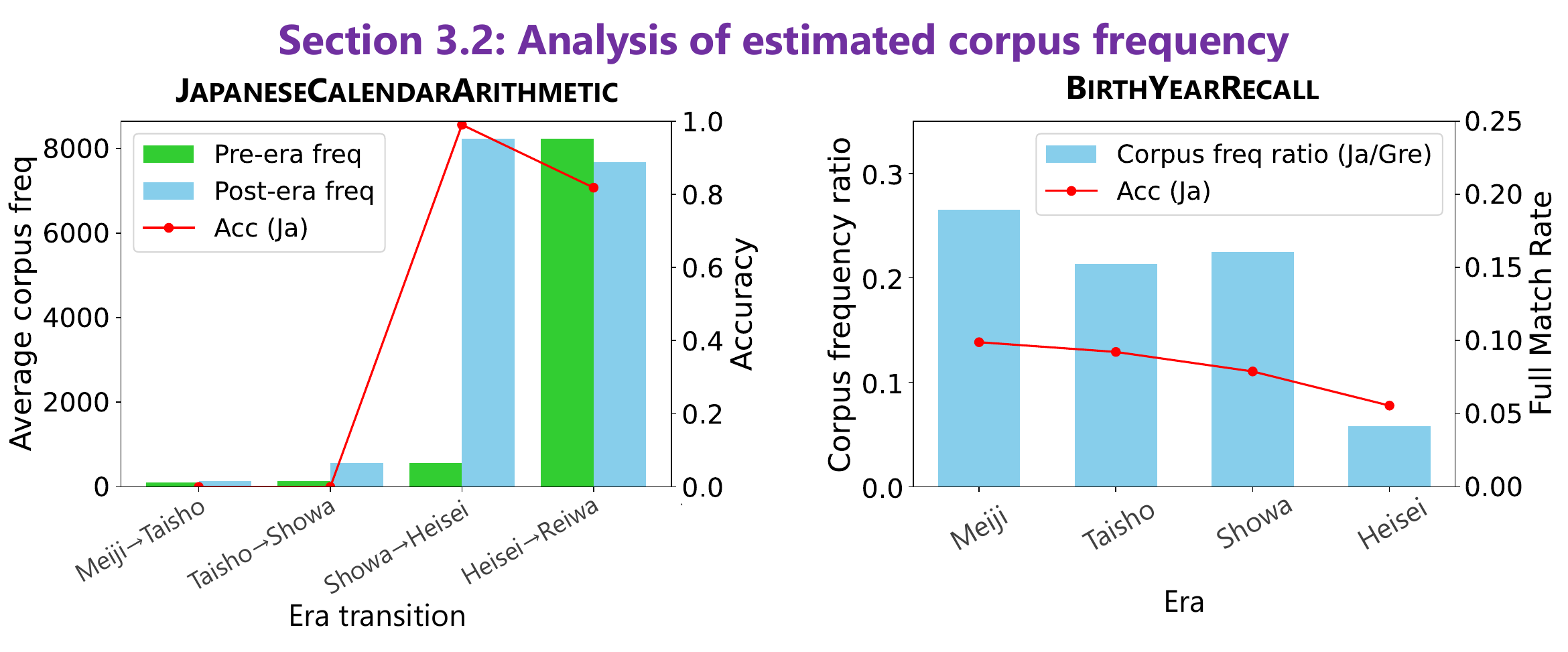}
\vspace{-4ex}
\caption{
Analysis of why many models fail from the perspective of estimated corpus frequency.
For llm-jp-3-13b, the only model with a public corpus, frequency analysis suggests that missing \wareki expressions cause failures in \textsc{\textsc{JapaneseCalendarArithmetic}}, while Gregorian bias explains lower accuracy in \textsc{BirthYearRecall}.
}
\label{fig:estimated_corpus_frequency}
\end{figure*}

\paragraph{Results.}
\cref{fig:human_recall} compares the accuracy of \textsc{BirthYearRecall} in Gregorian (x-axis) versus Japanese calendar years (y-axis).
The results indicate that all models exhibit a clear bias toward the Gregorian calendar, indicating that even Japanese-centric LMs mainly store birth years in the Gregorian format.
While English-centric LMs perform poorly on \wareki recall, Japanese-centric and frontier LMs show reasonable ability, though they still do better with the Gregorian calendar overall.

Moreover, evaluation using a more lenient metric (\cref{sec:birth_year_recall_detail}) suggests that models may roughly recall the birth year at the era level or that minor shifts arise during internal conversions from the Gregorian to the Japanese calendar.

Furthermore, we also examined the consistency of \textsc{BirthYearRecall} across different calendar systems (Japanese and Gregorian) by measuring the percentage of individuals for whom the model also correctly recalled the Gregorian year, given that it had already correctly recalled the \wareki birth year (\cref{sec:consistency}).
As a result, while some Japanese-centric models achieved over 80\% consistency, others remained around 50\%,
indicating that even among Japanese-centric models, there is substantial variation in how knowledge related to the Gregorian and Japanese calendars is recalled.

\section{Discussion}
\label{sec:discussion}
\subsection{Analysis of typical errors}
To gain an understanding of model failure modes, we analyzed typical errors.

In the \textsc{\textsc{JapaneseCalendarArithmetic}}, we examined the proportion of out-of-range errors (e.g., generating ``Heisei 37'' even though the Heisei era ended at year 31).
\cref{fig:typical_error} (top) shows that such an error was particularly pronounced during the Heisei→Reiwa transition across several LMs.
This result suggests that in later eras, out-of-range errors are one of the causes of failure.

For \textsc{BirthYearRecall}, we examined the proportion of Gregorian bias errors for each era.
This analysis focused on cases where models, despite being given three-shot \wareki examples designed to induce Japanese calendar responses, still produced answers in Gregorian years.
As shown in \cref{fig:typical_error} (bottom), this error is particularly pronounced in recent eras across all Japanese models, suggesting that Gregorian bias may hinder their ability to accurately recall years in \wareki format.

\subsection{Analysis of estimated corpus frequency}
Among the models investigated in this study, we examined llm-jp-3-13b, the only Japanese model with a publicly available pretraining corpus, trained on the llm-jp-corpus-v3~\citep{enomoto-etal-2024-investigating}. 
We used Infini-gram~\citep{Liu2024InfiniGram} to count year expressions in the pretraining corpus.

In \textsc{\textsc{JapaneseCalendarArithmetic}}, we analyzed the correlation between the frequency of \wareki year occurrences and the accuracy of this task.
We prepared Japanese expressions of \wareki ranging from Meiji 1 to Reiwa 11 and measured their frequencies in the corpus. 
For each era, we calculated the average frequency across its constituent years.
The results suggest a correlation between task accuracy and the corpus frequency of post-era Japanese calendar expressions (\cref{fig:estimated_corpus_frequency}, left).
In fact, the Pearson correlation coefficient between pre-era frequency and accuracy was 0.5086 (p = 0.4914), whereas that between post-era frequency and accuracy was much stronger, at 0.9959 (p = 0.0041).
This indicates that the poor performance likely stems from the underrepresentation of post-era expressions in the pretraining data.

In contrast, no positive correlation was observed between the frequency of \wareki year expressions and the accuracy in \textsc{BirthYearRecall}.
To investigate further, we measured the frequencies of both \wareki and corresponding Gregorian-year expressions in the corpus, averaged them within each era (from Meiji to Heisei), and calculated the ratio of Gregorian-year to \wareki expression frequencies.
This ratio showed a strong positive correlation with the accuracy of \textsc{BirthYearRecall} averaged over era (\cref{fig:estimated_corpus_frequency}, right), and the Pearson correlation was 0.9367 (p = 0.0633).
This suggests that in later eras, \wareki expressions appear relatively less frequently in the corpus than Gregorian ones, which may underlie the models' failures.

\section{Conclusions}
This work analyzed whether LMs can handle the Japanese calendar, a non-Gregorian calendar.
We evaluated models on tasks involving conversion, arithmetic, and factual recall.
While Japanese-centric LMs and frontier LMs handle basic conversions, most models struggle with more complex tasks and show inconsistent behavior.
Our findings highlight the need to understand the model limitations when dealing with non-Gregorian calendar systems and motivate future research investigating the causes of the revealed failures.

\clearpage
\section*{Limitations}
While our work reveals LMs have difficulty in handling the Japanese calendar, it has several limitations.

First, our study focuses only on the Japanese calendar. 
Among our three tasks, \textsc{CalendarConversion} and \textsc{BirthYearRecall} can be extended to other calendars using date pairs or biographical data. 
On the other hand, \textsc{JapaneseCalendarArithmetic} is specific to \wareki, where eras change irregularly with imperial succession. 
We believe that, when extending this line of research to other calendar systems, it is crucial to design similarly system-specific tasks that reflect each calendar system's unique temporal structure. 
For example, with the Hijri calendar, one could ask which Gregorian season Ramadan falls in for a given year. 
The task will test whether models understand how lunar cycles shift the timing of Ramadan across seasons.

Second, our evaluation relies on prompt-based testing and subsequent error analysis based on typical error patterns and corpus frequency, rather than directly examining the internal representations of calendrical knowledge.
While this analysis provides valuable insights into potential sources of error, it has limitations in identifying their causes in a direct, detailed manner.
Future work could employ probing or mechanistic interpretability methods to more precisely identify error sources and propose remedies.

Third, among currently available Japanese-centric models, llm-jp-3-13b is the only one with publicly released training data.
Therefore, the distribution of calendar-related expressions in other models' training data remains unknown.
The release of pretraining data from more Japanese-centric models may help explain performance differences across models.

\section*{Ethical Considerations}
All data created and/or used in this work was synthetically generated and/or derived from Wikidata, a public knowledge base released under the CC0~1.0 Universal license.
As such, we do not foresee any ethical concerns regarding personally identifying information or offensive content.
Also, the llm-jp-corpus-v3~\citep{enomoto-etal-2024-investigating} used in \cref{sec:discussion} is publicly available.

All language models used in this study are publicly available. 
We strictly adhered to the terms and conditions of each model's license, including, but not limited to, those released under the Meta Llama and Mistral licensing terms.

During the development of code and the writing of this paper, we made use of AI assistants, including large language models. 
All code snippets and textual content generated with the assistance of such tools were carefully reviewed and revised by the authors to ensure scientific integrity, accuracy, and ethical compliance.
\section*{Acknowledgements}
This work was supported by JST BOOST Grant Numbers JPMJBS2412, JPMJBS2421, JPMJBY24F9; JST CREST Grant Number JPMJCR20D2; JSPS KAKENHI Grant Number JP25K03175; AMED Grant Number JP25wm0625405, and the Nakajima Foundation.

\bibliography{custom}

\clearpage
\onecolumn
\appendix
\label{sec:appendix}
\section{Datasets}\label{sec:datasets}
The datasets used in each task are presented in \cref{tab:dataset_for_calendar_conversion} for \textsc{CalendarConversion}, \cref{tab:dataset_for_Japanese_calendar_arithmetic_after10} and \cref{tab:dataset_for_Japanese_calendar_arithmetic_before10} for \textsc{JapaneseCalendarArithmetic}, and \cref{tab:dataset_for_birth_year_recall} for \textsc{BirthYearRecall}.

For \textsc{CalendarConversion}, we constructed a dataset of corresponding Gregorian and \wareki years for the Meiji (1868–1912), Taisho (1912–1926), Showa (1926–1989), and Heisei (1989–2019) eras.

For \textsc{JapaneseCalendarArithmetic}, we sampled 500 unique and non-overlapping dates for each era. 
In the after-ten-years setting, dates were randomly selected from the last five years of the Meiji, Taisho, Showa, and Heisei eras. 
In the before-ten-years setting, dates were taken from the first five years of the Taisho, Showa, Heisei, and Reiwa eras.
To ensure temporal precision at era boundaries, we carefully constructed the dataset such that, for example, Heisei dates end on April 30, Heisei 31, and Reiwa dates begin on May 1, Reiwa 1. 
This ensures that the model must reason across era boundaries to generate a correct answer.

\begin{figure*}
\centering
\includegraphics[width=\linewidth]{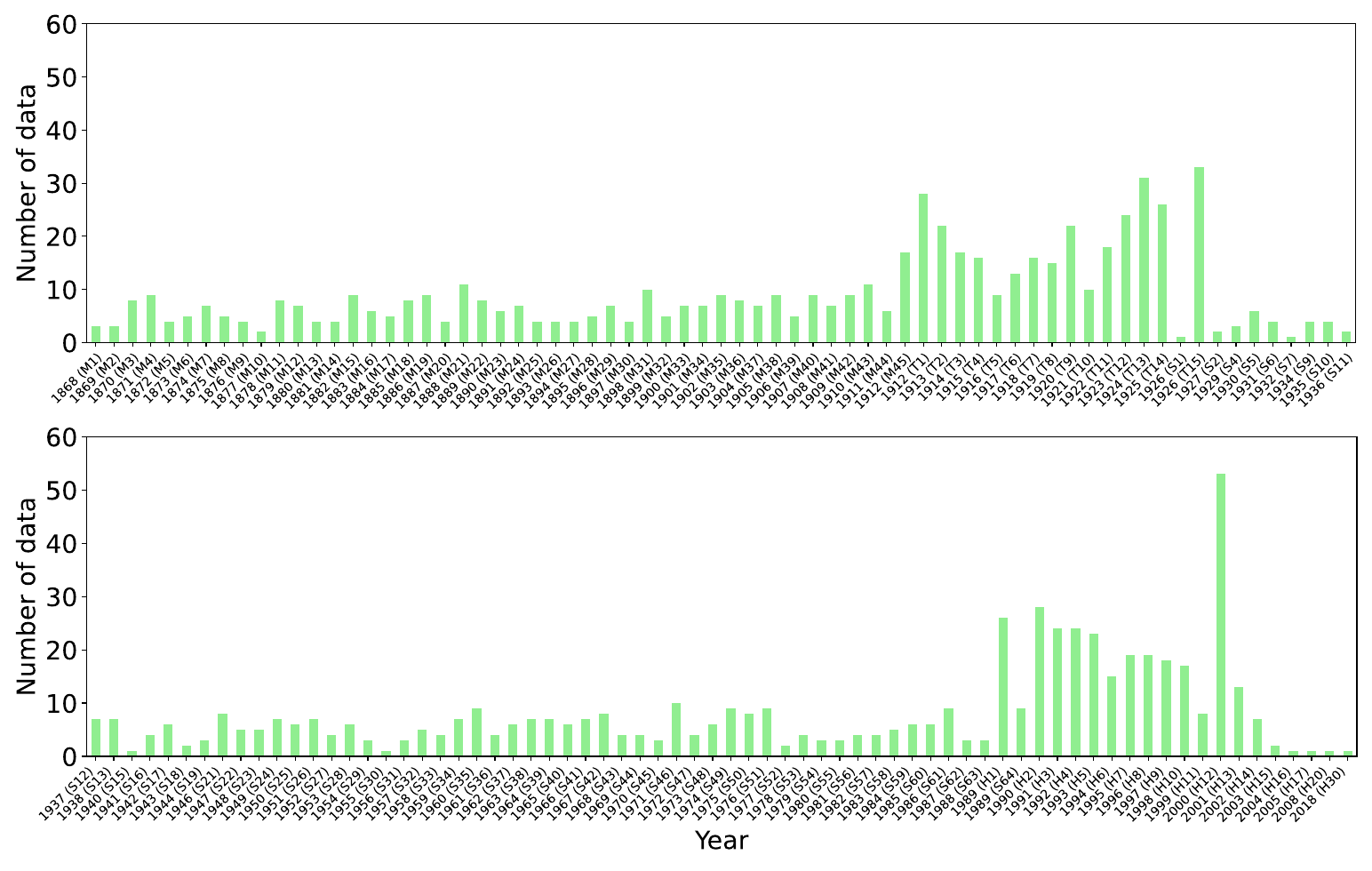}
\caption{
The distribution of birth-year data used in \textsc{BirthYearRecall}.
The horizontal axis represents the year, and the vertical axis represents the number of data samples.
The letters M, T, S, and H on the horizontal axis correspond to Meiji, Taisho, Showa, and Heisei, respectively.
Years without bars indicate that there are zero samples for individuals born in that year.
As mentioned in \cref{sec:birth_year_recall}, a total of 300 individuals were sampled for each era, resulting in 1,200 data samples in total.
}
\label{fig:birth_year_frequency}
\end{figure*}

For \textsc{BirthYearRecall}, we extracted 300 Japanese individuals per era from Wikidata, filtering for entities with at least 20 relations. 
In total, 1,200 individuals were sampled across the Meiji, Taisho, Showa, and Heisei eras. 
The distribution of birth year data for the individuals is shown in \cref{fig:birth_year_frequency}.
Each entry includes the individual's name and birth year in both Japanese and English formats to accommodate prompts in both languages.

\begin{table*}[t]
    \centering
    \begin{tabular}{ccccc}
        \toprule
        Era & Lang & \# & Gregorian & Japanese\\
        \midrule
        Meiji & en & 1 &1868 & Meiji 1 \\
        Meiji & en & 2 &1869 & Meiji 2 \\
        \vdots & \vdots & \vdots & \vdots &\vdots \\
        Meiji & en & 45 &1912 & Meiji 45 \\
        \midrule[0.01em]
        Meiji & ja & 1 & 1868年 & 明治1年 \\
        Meiji & ja & 2 & 1869年 & 明治2年 \\
        \vdots & \vdots & \vdots & \vdots &\vdots \\
        Meiji & ja & 45 & 1912年 & 明治45年 \\
        \midrule[0.01em]
        Taisho & en & 1 & 1912 & Taisho 1 \\
        Taisho & en & 2 & 1913 & Taisho 2 \\
        \vdots & \vdots & \vdots & \vdots &\vdots \\
        Taisho & en & 15 & 1926 & Taisho 15 \\
        \midrule[0.01em]
        Taisho & ja & 1 & 1912年 & 大正1年 \\
        Taisho & ja & 2 & 1913年 & 大正2年 \\
        \vdots & \vdots & \vdots & \vdots &\vdots \\
        Taisho & en & 15 & 1926年 & 大正15年 \\
        \midrule[0.01em]
        Showa & en & 1 & 1926 & Showa 1 \\
        Showa & en & 2 & 1927 & Showa 2 \\
        \vdots & \vdots & \vdots & \vdots &\vdots \\
        Showa & en & 64 & 1989 & Showa 64 \\
        \midrule[0.01em]
        Showa & ja & 1 & 1926年 & 昭和1年 \\
        Showa & ja & 2 & 1927年 & 昭和2年 \\
        \vdots & \vdots & \vdots & \vdots &\vdots \\
        Showa & ja & 64 & 1989年 & 昭和64年 \\
        \midrule[0.01em]
        Heisei & en & 1 & 1989 & Heisei 1 \\
        Heisei & en & 2 & 1990 & Heisei 2 \\
        \vdots & \vdots & \vdots & \vdots &\vdots \\
        Heisei & en & 31 & 2019 & Heisei 31 \\
        \midrule[0.01em]
        Heisei & ja & 1 & 1989年 & 平成1年 \\
        Heisei & ja & 2 & 1990年 & 平成2年 \\
        \vdots & \vdots & \vdots & \vdots &\vdots \\
        Heisei & ja & 31 & 2019年 & 平成31年 \\
        \bottomrule
    \end{tabular}
    \caption{Dataset used for \textsc{CalendarConversion}}
    \label{tab:dataset_for_calendar_conversion}
\end{table*}

\begin{table*}[t]
\centering
\begin{tabular}{ccccc}
\toprule
Era & Lang & \# & Date (Japanese calendar) & Gold date (Japanese calendar)  \\
\midrule
Meiji & en & 1 & August 24, Meiji 41 & August 24, Taisho 7 \\
Meiji & en & 2 & April 30, Meiji 42 & April 30, Taisho 8 \\
\vdots & \vdots & \vdots & \vdots &\vdots \\
Meiji & en & 500 & August 1, Meiji 42 & August 1, Taisho 8 \\
\midrule[0.01em]
Meiji & ja & 1 & 明治41年8月24日 & 大正7年8月24日 \\
Meiji & ja & 2 & 明治42年4月30日 & 大正8年4月30日 \\
\vdots & \vdots & \vdots & \vdots &\vdots \\
Meiji & ja & 500 & 明治42年8月1日 & 大正8年8月1日 \\
\midrule[0.01em]
Taisho & en & 1 & November 13, Taisho 12 & November 13, Showa 8 \\
Taisho & en & 2 & February 5, Taisho 12 & February 5, Showa 8 \\
\vdots & \vdots & \vdots & \vdots &\vdots \\
Taisho & en & 500 & November 23, Taisho 15 & November 23, Showa 11 \\
\midrule[0.01em]
Taisho & ja & 1 & 大正12年11月13日 & 昭和8年11月13日 \\
Taisho & ja & 2 & 大正12年2月5日 & 昭和8年2月5日 \\
\vdots & \vdots & \vdots & \vdots &\vdots \\
Taisho & ja & 500 & 大正15年11月23日 & 昭和11年11月23日 \\
\midrule[0.01em]
Showa & en & 1 & December 11, Showa 63 & December 11, Heisei 10 \\
Showa & en & 2 & September 22, Showa 62 & September 22, Heisei 9 \\
\vdots & \vdots & \vdots & \vdots &\vdots \\
Showa & en & 500 & April 7, Showa 63 & April 7, Heisei 10 \\
\midrule[0.01em]
Showa & ja & 1 & 昭和63年12月11日 & 平成10年12月11日 \\
Showa & ja & 2 & 昭和62年9月22日 & 平成9年9月22日 \\
\vdots & \vdots & \vdots & \vdots &\vdots \\
Showa & ja & 500 & 昭和63年4月7日 & 平成10年4月7日 \\
\midrule[0.01em]
Heisei & en & 1 & November 9, Heisei 28 & November 9, Reiwa 8 \\
Heisei & en & 2 & December 24, Heisei 30 & December 24, Reiwa 10 \\
\vdots & \vdots & \vdots & \vdots &\vdots \\
Heisei & en & 500 & February 27, Heisei 31 & February 27, Reiwa 11 \\
\midrule[0.01em]
Heisei & ja & 1 & 平成28年11月9日 & 令和8年11月9日 \\
Heisei & ja & 2 & 平成30年12月24日 & 令和10年12月24日 \\
\vdots & \vdots & \vdots & \vdots &\vdots \\
Heisei & ja & 500 & 平成31年2月27日 & 令和11年2月27日 \\
\bottomrule
\end{tabular}
\caption{Dataset used for \textsc{JapaneseCalendarArithmetic} (add ten years)}
\label{tab:dataset_for_Japanese_calendar_arithmetic_after10}
\end{table*}

\begin{table*}[t]
    \centering
    \begin{tabular}{ccccc}
        \toprule
        Era & Lang & \# & Date (Japanese calendar) & Gold date (Japanese calendar)  \\
        \midrule
        Taisho & en & 1 & August 28, Taisho 5 & August 28, Meiji 39 \\
        Taisho & en & 2 & December 22, Taisho 4 & December 22, Meiji 38 \\
        \vdots & \vdots & \vdots & \vdots &\vdots \\
        Taisho & en & 500 & September 20, Taisho 4 & September 20, Meiji 38 \\
        \midrule[0.01em]
        Taisho & ja & 1 & 大正5年8月28日 & 明治39年8月28日 \\
        Taisho & ja & 2 & 大正4年12月22日 & 明治38年12月22日\\
        \vdots & \vdots & \vdots & \vdots &\vdots \\
        Taisho & ja & 500 & 大正4年9月20日 & 明治38年9月20日 \\
        \midrule[0.01em]
        Showa & en & 1 & January 21, Showa 6 & January 21, Taisho 10 \\
        Showa & en & 2 & October 29, Showa 6 & October 29, Taisho 10 \\
        \vdots & \vdots & \vdots & \vdots &\vdots \\
        Showa & en & 500 & February 19, Showa 2 & February 19, Taisho 6 \\
        \midrule[0.01em]
        Showa & ja & 1 & 昭和6年1月21日 & 大正10年1月21日 \\
        Showa & ja & 2 & 昭和6年10月29日 & 大正10年10月29日 \\
        \vdots & \vdots & \vdots & \vdots &\vdots \\
        Showa & ja & 500 & 昭和2年2月19日 & 大正6年2月19日 \\
        \midrule[0.01em]
        Heisei & en & 1 & January 25, Heisei 1 & January 25, Showa 54\\
        Heisei & en & 2 & September 4, Heisei 1 & September 4, Showa 54 \\
        \vdots & \vdots & \vdots & \vdots &\vdots \\
        Heisei & en & 500 & May 29, Heisei 1 & May 29, Showa 54 \\
        \midrule[0.01em]
        Heisei & ja & 1 & 平成1年1月25日 & 昭和54年1月25日 \\
        Heisei & ja & 2 & 平成1年9月4日 & 昭和54年9月4日 \\
        \multicolumn{1}{c}{\ldots} & \multicolumn{1}{c}{\ldots} & \multicolumn{1}{c}{\ldots} & 
        \multicolumn{1}{c}{\ldots} & \multicolumn{1}{c}{\ldots} \\
        Heisei & ja & 500 & 平成1年5月29日 & 昭和54年5月29日 \\
        \midrule[0.01em]
        Reiwa & en & 1 & July 7, Reiwa 4 & July 7, Heisei 24 \\
        Reiwa & en & 2 & September 19, Reiwa 3 & September 19, Heisei 23 \\
        \vdots & \vdots & \vdots & \vdots &\vdots \\
        Reiwa & en & 500 & July 4, Reiwa 5 & July 4, Heisei 25 \\
        \midrule[0.01em]
        Reiwa & ja & 1 & 令和4年7月7日 & 平成24年7月7日 \\
        Reiwa & ja & 2 & 令和3年9月19日 & 平成23年9月19日 \\
        \vdots & \vdots & \vdots & \vdots &\vdots \\
        Reiwa & ja & 500 & 令和5年7月4日 & 平成25年7月4日 \\
        \bottomrule
    \end{tabular}
    \caption{Dataset used for \textsc{JapaneseCalendarArithmetic} (subtract ten years)}
    \label{tab:dataset_for_Japanese_calendar_arithmetic_before10}
\end{table*}

\begin{table*}[t]
    \centering
    \tabcolsep 3pt
    \begin{tabular}{cccccc}
        \toprule
        Era & Lang & \# & Name & Gold birth year (Gregorian) & Gold birth year (Japanese) \\
        \midrule
        Meiji & en & 1 & Bunji Tsushima & 1898 & Meiji 31 \\
        Meiji & en & 2 & Heinosuke Gosho & 1902 & Meiji 35 \\
        \vdots & \vdots & \vdots & \vdots &\vdots&\vdots \\
        Meiji & en & 300 & Kikuko, Princess Takamatsu & 1911 & Meiji 44 \\ 
        \midrule[0.01em]
        Meiji & ja & 1 & 津島文治 & 1898年 & 明治31年 \\
        Meiji & ja & 2 & 五所平之助 & 1902年 & 明治35年 \\
        \vdots & \vdots & \vdots & \vdots &\vdots&\vdots \\
        Meiji & ja & 300 & 宣仁親王妃喜久子 & 1911年 & 明治44年 \\ 
        \midrule[0.01em]
        Taisho & en & 1 & Tetsuo Takaha & 1926 & Taisho 15 \\
        Taisho & en & 2 & Kiyoshi Ito & 1915 & Taisho 4 \\
        \vdots & \vdots & \vdots & \vdots &\vdots &\vdots\\
        Taisho & en & 300 & Yozo Matsushima & 1921 & Taisho 10 \\ 
        \midrule[0.01em]
        Taisho & ja & 1 & 高羽哲夫 & 1926年 & 大正15年 \\
        Taisho & ja & 2 & 伊藤清 & 1915年 & 大正4年 \\
        \vdots & \vdots & \vdots & \vdots &\vdots&\vdots \\
        Taisho & ja & 300 & 松島与三 & 1921年 & 大正10年 \\ 
        \midrule[0.01em]
        Showa & en & 1 & Hiroshi Katsuno & 1949 & Showa 24 \\
        Showa & en & 2 & Homare Sawa & 1978 & Showa 53 \\
        \vdots & \vdots & \vdots & \vdots &\vdots &\vdots\\
        Showa & en & 300 & Hideki Matsui & 1974 & Showa 49 \\ 
        \midrule[0.01em]
        Showa & ja & 1 & 勝野洋 & 1949年 & 昭和24年 \\
        Showa & ja & 2 & 澤穂希 & 1978年 & 昭和53年 \\
        \vdots & \vdots & \vdots & \vdots &\vdots &\vdots\\
        Showa & ja & 300 & 松井秀喜 & 1974年 & 昭和49年 \\ 
        \midrule[0.01em]
        Heisei & en & 1 & Miyuri Shimabukuro & 1994 & Heisei 6 \\
        Heisei & en & 2 & Sakura Miyawaki & 1998 & Heisei 10 \\
        \vdots & \vdots & \vdots & \vdots &\vdots &\vdots\\
        Heisei & en & 300 & Maimi Yajima & 1992 & Heisei 4 \\ 
        \midrule[0.01em]
        Heisei & ja & 1 & 島袋美由利 & 1994年 & 平成6年 \\
        Heisei & ja & 2 & 宮脇咲良 & 1998年 & 平成10年 \\
        \vdots & \vdots & \vdots & \vdots &\vdots&\vdots \\
        Heisei & ja & 300 & 矢島舞美 & 1992年 & 平成4年 \\ 
        \bottomrule
    \end{tabular}
    \caption{Dataset used for \textsc{BirthYearRecall}}
    \label{tab:dataset_for_birth_year_recall}
\end{table*}

\section{Prompts}\label{sec:prompts}
The prompts used for the four tasks in our experiments are summarized in \cref{tab:prompt}. 
For each task, we prepared both English and Japanese versions of the prompts. 
English prompts were used with English-centric models, and Japanese prompts were used with Japanese-centric models.
In frontier models (GPT-4o and DeepSeekV3), since these models can handle both Japanese and English prompts, we use both Japanese and English prompts.
To ensure consistency, we used the same user prompts as for the other models.
Also, to induce responses in the intended format, we set the system prompt as follows:

Japanese: "あなたは和暦とグレゴリオ暦の専門家です。以下に続くように文章を答えのみ生成してください。"

English: "You are an expert in the Japanese and Gregorian calendars. Please generate only the answer that continues from the text below."

For \textsc{CalendarConversion}, the prompts request conversion between Gregorian and \wareki dates, in both directions. 
A one-shot example was provided before the prompt to ensure that the model outputs the answer in the correct format, either as a four-digit year for Gregorian dates or as a combination of an era name and a year for Japanese dates.

For \textsc{JapaneseCalendarArithmetic}, the prompts ask about the year ten years before or after a given date, often across era boundaries. 
As with the previous task, a one-shot example was shown in advance to guide the model toward the correct output format (e.g., “August 29, Reiwa 10”).

For \textsc{BirthYearRecall}, the prompts asked for the birth year of a Japanese individual in either the Gregorian calendar or the \wareki.
We provided three-shot examples before the prompt to help the model produce answers in the correct format, either as a four-digit number for Gregorian dates or as an era name followed by a year for Japanese dates.

\begin{table*}[t]
    \centering
    \small
    \tabcolsep 3pt
    \begin{tabular}{cccccc}
        \toprule
        Task Type & Lang & Option & \# & Prompt(example)\\
        \midrule
        \textsc{CalendarConversion} & en & GtoJ & 1 & In the Japanese calendar, the year 1992 corresponds to\\
        \textsc{CalendarConversion} & en & JtoG & 1 & In the Gregorian calendar, Heisei 4 corresponds to the year \\
        \textsc{CalendarConversion} & ja & GtoJ & 1 & 平成4年を西暦に変換すると、\\
        \textsc{CalendarConversion} & ja & JtoG & 1 & 1992年を和暦に変換すると、\\
        \midrule[0.01em]
        \textsc{JapaneseCalendarArithmetic} & en & +10yr & 1 & Ten years after August 29, Heisei 30 is \\
        \textsc{JapaneseCalendarArithmetic} & en & +10yr & 2 & If you go forward 10 years from August 29, Heisei 30, you get \\
        \textsc{JapaneseCalendarArithmetic} & en & +10yr & 3 & The date 10 years after August 29, Heisei 30 is \\
        \textsc{JapaneseCalendarArithmetic} & en & -10yr & 1 & Ten years before April 27, Heisei 3 is \\
        \textsc{JapaneseCalendarArithmetic} & en & -10yr & 2 & If you go back 10 years from April 27, Heisei 3, you get \\
        \textsc{JapaneseCalendarArithmetic} & en & -10yr & 3 & The date 10 years prior to April 27, Heisei 3 is \\
        \textsc{JapaneseCalendarArithmetic} & ja & +10yr & 1 & 平成30年8月29日の10年後は \\
        \textsc{JapaneseCalendarArithmetic} & ja & +10yr & 2 & 平成30年8月29日から10年経つと \\
        \textsc{JapaneseCalendarArithmetic} & ja & +10yr & 3 & 平成30年8月29日に対する10年後の日付は \\
        \textsc{JapaneseCalendarArithmetic} & ja & -10yr & 1 & 平成3年4月27日の10年前は \\
        \textsc{JapaneseCalendarArithmetic} & ja & -10yr & 2 & 平成3年4月27日から10年さかのぼると \\
        \textsc{JapaneseCalendarArithmetic} & ja & -10yr & 3 & 平成3年4月27日に至る10年前の日付は \\
        \midrule[0.01em]
        \textsc{BirthYearRecall}  & en & G & 1 & According to the Gregorian calendar, Hideki Matsui was born in \\
        \textsc{BirthYearRecall}  & en & G & 2 & The Gregorian calendar states that Hideki Matsui was born in \\
        \textsc{BirthYearRecall}  & en & G & 3 & The Gregorian calendar dates Hideki Matsui's birth to \\
        \textsc{BirthYearRecall}  & en & J & 1 & According to the Japanese calendar, Hideki Matsui was born in \\
        \textsc{BirthYearRecall}  & en & J & 2 & The Japanese calendar states that Hideki Matsui was born in \\
        \textsc{BirthYearRecall}  & en & J & 3 & The Japanese calendar dates Hideki Matsui's birth to \\
        \textsc{BirthYearRecall}  & ja & G & 1 & 西暦で松井秀喜が生まれたのは \\
        \textsc{BirthYearRecall}  & ja & G & 2 & 松井秀喜の誕生年は \\
        \textsc{BirthYearRecall}  & ja & G & 3 & 松井秀喜の生まれ年は \\
        \textsc{BirthYearRecall}  & ja & J & 1 & 和暦で松井秀喜が生まれたのは \\
        \textsc{BirthYearRecall}  & ja & J & 2 & 松井秀喜の誕生年は \\
        \textsc{BirthYearRecall}  & ja & J & 3 & 松井秀喜の生まれ年は \\
        \bottomrule
    \end{tabular}
    \caption{
    Prompts used for each task. 
    In the Option column, GtoJ and JtoG denote Gregorian to Japanese calendar and Japanese to Gregorian \textsc{CalendarConversion}, respectively. 
    +10yr and -10yr indicate addition and subtraction of ten years in \textsc{JapaneseCalendarArithmetic}. 
    G and J represent prompts requiring output in the Gregorian and Japanese calendars, respectively.
    Few-shot examples were added before each prompt to guide the model to respond in the correct format.
    }
    \label{tab:prompt}
\end{table*}

\begin{table*}[t]
    \centering
    \small
    \begin{tabular}{lccc}
        \toprule
        Model name & Lang & paper & Repo name on Huggingface \\
        \midrule
        Llama-2-7b & en & \citet{touvron2023llama2openfoundation} & meta-llama/Llama-2-7b \\
        Llama-2-13b & en & \citet{touvron2023llama2openfoundation} & meta-llama/Llama-2-13b \\
        Mistral-7B & en & \citet{jiang2023mistral7b} & mistralai/Mistral-7B-v0.1 \\
        Llama3.1-8B & en & \citet{grattafiori2024Llama3herdmodels} & meta-llama/Llama-3.1-8B \\
        llm-jp-3-13b & ja & \citet{llmjp2024llmjpcrossorganizationalprojectresearch} & llm-jp/llm-jp-3-13b\\
        sarashina2-13b & ja & - & sbintuitions/sarashina2-13b \\
        Swallow-13b & ja & \citet{swallow_01, swallow_02} & tokyotech-llm/Swallow-13b-hf \\
        Swallow-MS-7b & ja & - & tokyotech-llm/Swallow-MS-7b-v0.1 \\
        Llama3-Swallow-8B & ja & - & tokyotech-llm/Llama-3-Swallow-8B-v0.1 \\
        \bottomrule
    \end{tabular}
    \caption{List of Japanese-centric and English-centric models used in the experiments.}
    \label{tab:modellist}
\end{table*}

\section{Models}\label{sec:models}
We use four English-centric models: Llama-2-7B, Llama-2-13B~\citep{touvron2023llama2openfoundation}, Mistral-7B~\citep{jiang2023mistral7b}, and Llama-3.1-8B~\citep{grattafiori2024Llama3herdmodels}.
For Japanese-centric models, we include two trained from scratch in Japanese (llm-jp-3-13b~\citep{llmjp2024llmjpcrossorganizationalprojectresearch} and sarashina2-13b), and three models continued pretraining in Japanese: Swallow-13b~\citep{swallow_01, swallow_02}, Swallow-MS-7b, and Llama3-Swallow-8B.
All experiments were conducted using a single RTX 6000 Ada (48GB) GPU.
Also, we used GPT-4o~\citep{openai2024gpt4ocard} and DeepSeek V3~\citep{deepseekai2025deepseekv3technicalreport} as comparative baselines for the frontier models.

\clearpage
\section{\textsc{CalendarConversion}}
\begin{figure*}
\centering
\includegraphics[width=\linewidth]{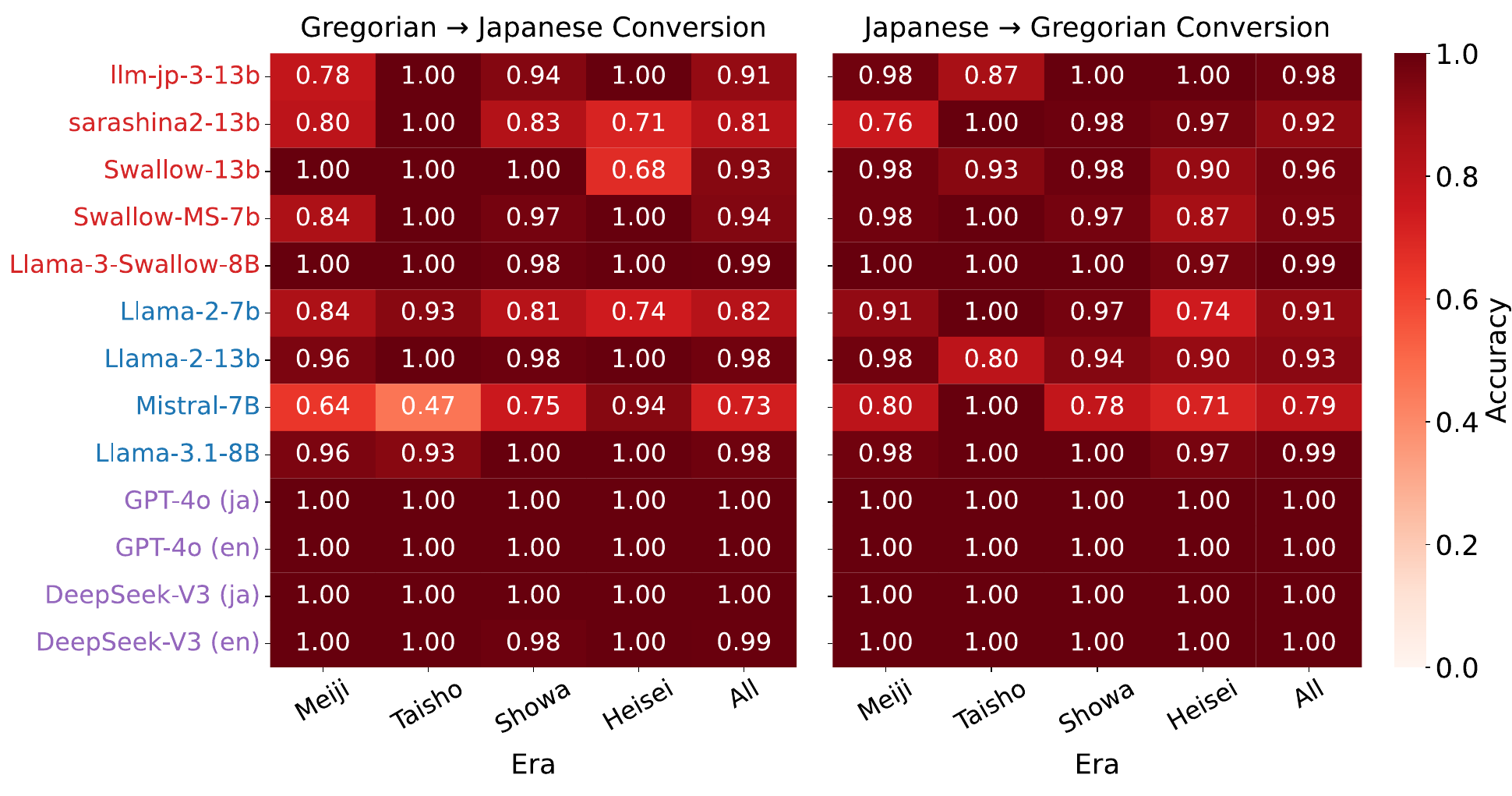}
\caption{
Accuracy of Gregorian-to-Japanese (left) and Japanese-to-Gregorian (right) \textsc{CalendarConversion}.
Japanese-centric models and frontier models achieved near-perfect accuracy in both directions, while English-centric models showed notable variation across models and eras.
}
\label{fig:calendar_conversion_detail}
\end{figure*}

\cref{fig:calendar_conversion_detail} shows the results of \textsc{CalendarConversion}: from Gregorian to \wareki (left) and from \wareki to Gregorian calendar (right).
While Japanese-centric models, GPT-4o, and DeepSeek V3 perform near-perfect conversions in both directions, English-centric models exhibit greater variance across models and eras, generally showing inferior performance.
Nevertheless, some English models, such as Llama3.1-8B, demonstrate high accuracy in simple conversions.
In certain cases, such as Mistral-7B for the Taishō era, models succeed in only one conversion direction, highlighting asymmetries in their learned representations.

\section{\textsc{JapaneseCalendarArithmetic}}\label{sec:appendix_arithmetic}
To show the details of the results in \textsc{JapaneseCalendarArithmetic}, we introduce three metrics: Era Match, Near Match, and Full Match.

\textbf{Era Match} captures a coarse understanding of Japanese calendar eras and is defined as the ratio of outputs containing the correct era:$\frac{1}{N} \sum_{i=1}^N \mathbbm{1}(\hat{E}_i = E_i)$.
In the above example, ``September 19, Heisei 37'' would be incorrect, as the Heisei era ended before reaching year 37, while ``September 20, Reiwa 6'' is a correct era match.

\textbf{Near Match} accounts for the difficulty of conversions involving transition years and is defined as the ratio of predictions that are off by at most one year:$\frac{1}{N} \sum_{i=1}^N \mathbbm{1}(|\mathrm{G}(E_i, x_i) - \mathrm{G}(\hat{E}_i, \hat{x}_i)| \leq 1)$, where $\mathrm{G}(E, x)$ converts a \wareki era and year to its Gregorian equivalent.  

\textbf{Full Match} jointly measures knowledge of era transitions and year arithmetic. 
It is defined as the ratio of outputs that exactly match the correct date:$\frac{1}{N} \sum_{i=1}^N \mathbbm{1}(E_i = \hat{E}_i \land x_i = \hat{x}_i)$.
This metric is the same as the accuracy used in \cref{sec: Japanese_calendar_arithmetic}.

\begin{figure*}
\centering
\includegraphics[width=\linewidth]{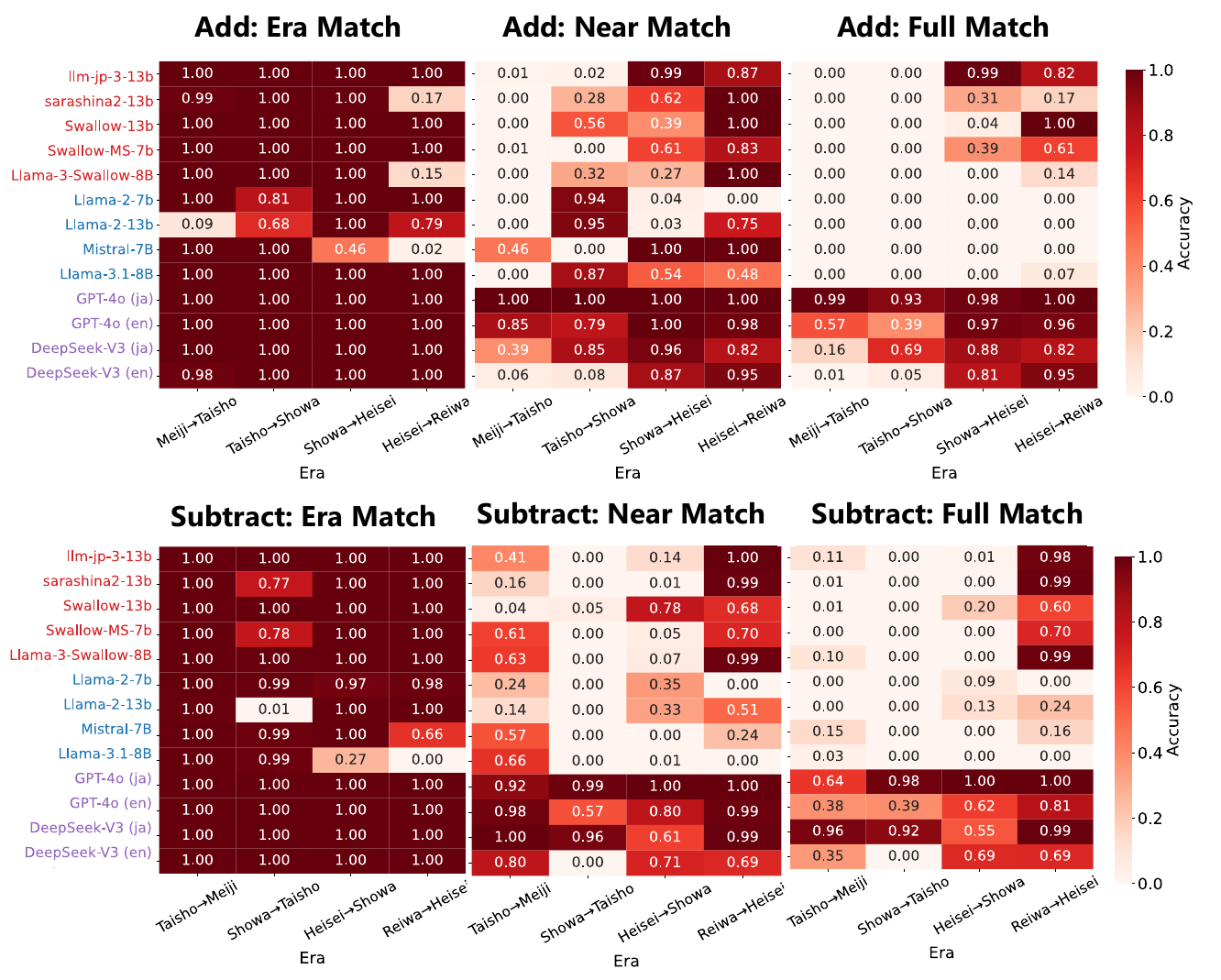}
\caption{
Evaluation results for three metrics(Era match Accuracy, Near match Accuracy, and Full match Accuracy) in \textsc{JapaneseCalendarArithmetic}: 10-year addition (top) and 10-year subtraction (bottom).
Both Japanese-centric and English-centric models achieve high Era match accuracies for most transitions, indicating that they generally understand the chronological order of era.
However, for the Full match accuracy, only Japanese models show high performance on recent transitions, suggesting that they capture the timeline discontinuity at era boundaries.
English models, by contrast, fail to do so, likely because they do not recognize that era transitions (e.g., Heisei 31 to Reiwa 1) can occur within the same Gregorian year.
}

\label{fig:arithmetic_all}
\end{figure*}

The results are shown in \cref{fig:arithmetic_all}.
The top half shows the results for adding ten years, and the bottom half shows the results for subtracting ten years, across Japanese era boundaries.

From the Era Match accuracy, we can see that most models seem to understand the basic order of the eras, with a few exceptions.
However, as discussed in \cref{sec: Japanese_calendar_arithmetic}, the Full Match accuracy shows that even Japanese-centric models and frontier models consistently fail to reason correctly across transitions between older eras, such as Meiji to Taisho or Taisho to Showa.
In contrast, for more recent transitions, such as Heisei to Reiwa, Japanese-centric models perform much better than English-centric models.

Japanese models and frontier models usually get higher Full Match accuracies for newer eras, but English models still show low accuracy even in those cases.
Many models, especially English ones, show a big gap between their Near Match accuracy and Full Match accuracy.
This means they often give answers that are just one year off and cannot handle era transitions exactly.
One main reason for these mistakes is that the models do not take into account that era transitions often happen in the same Gregorian year (e.g., Heisei 31 and Reiwa 1 both correspond to 2019).

To sum up, most Japanese-centric models seem to understand the timeline of recent eras well enough to reason correctly across era boundaries, whereas English-centric models still struggle with this.

\section{\textsc{BirthYearRecall}}
To show the details of the results in \textsc{BirthYearRecall}, we introduce two metrics: Full match and Within ±3 years Match.
\label{sec:birth_year_recall_detail}
\textbf{Full Match} requires an exact match of the prediction and the target.
For Gregorian output, accuracy is
$\frac{1}{N} \sum_{i=1}^N \mathbbm{1}(\hat{y}_i = y_i)$.
For \wareki output, the prediction must match both the era and the year within the era:
$\frac{1}{N} \sum_{i=1}^N \mathbbm{1}(E_i = \hat{E}_i \land x_i = \hat{x}_i)$.
This metric is the same as the accuracy used in \cref{sec:birth_year_recall}

\textbf{Within ±3 years Match} allows a deviation of $\pm$3 years.  
For \wareki, this is defined as:
$\frac{1}{N} \sum_{i=1}^N \mathbbm{1}(E_i = \hat{E}_i \land |x_i - \hat{x}_i| \leq 3)$
meaning that the prediction must be in the same era and within a 3-year range.
In the Gregorian setting, we convert the predicted and target year into \wareki values $(E, x)$ and then apply the tolerance match condition.

\begin{figure*}
\centering
\includegraphics[width=0.6\linewidth]{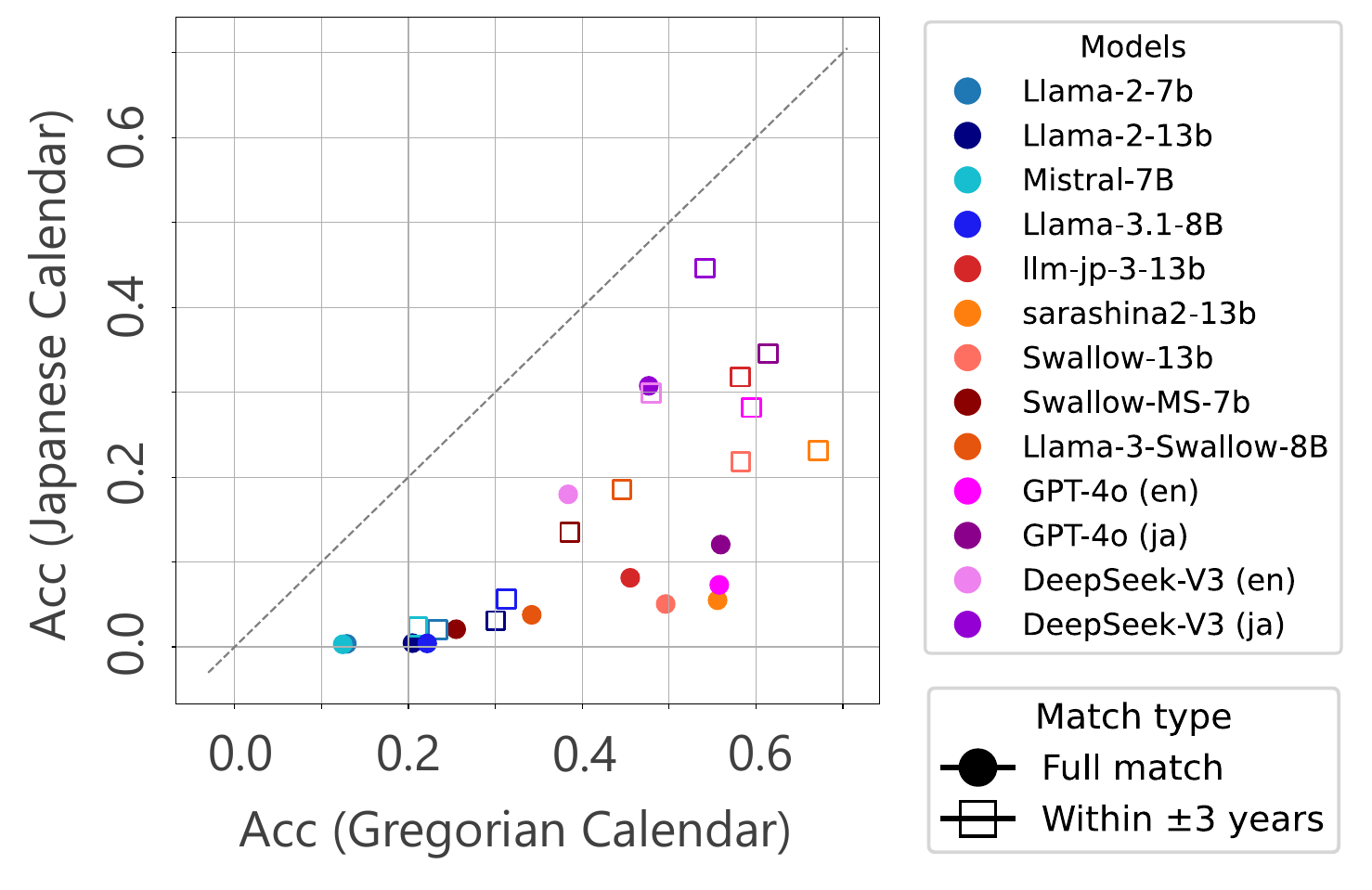}
\caption{
Results of \textsc{BirthYearRecall} using both Japanese and English prompts.
The accuracy of both Japanese-centric and English-centric models varies significantly depending on whether the person's name is presented in Japanese or English.
For Japanese calendar outputs, English models fail to produce correct answers regardless of the name format, while Japanese models only succeed when the name is presented in Japanese.
}
\label{fig:human_recall_detail}
\end{figure*}

\cref{fig:human_recall_detail} shows the results of the two metrics in \textsc{BirthYearRecall}. 
The x-axis indicates accuracy in the Gregorian calendar, and the y-axis indicates accuracy in 
\wareki.

English-centric models perform poorly in recalling birth years in \wareki. 
Japanese-centric models and frontier LMs show moderate success when prompted in Japanese, but still underperform compared to their accuracy in Gregorian date recall. 
Although Japanese-centric models are trained on Japanese corpora, they mainly store birth years in the Gregorian format.

For all Japanese LMs, the Within ±3-years match accuracy for \wareki recall is more than three times higher than the Full Match accuracy.
This suggests that models may either retrieve era-based years with some inaccuracy or rely on internal Gregorian-to-Japanese conversions that result in small shifts.

\section{\textsc{CrossCalendarConsistency}}
\label{sec:consistency}
\paragraph{Settings.}
This task evaluates the consistency of \textsc{BirthYearRecall} across calendars.
Specifically, it measures the ratio of individuals for whom the model correctly predicts the birth year both in \wareki and in the Gregorian calendar.
We define \textbf{Full Match Consistency} as:
$\frac{1}{M} \sum_{i=1}^M \mathbbm{1}(\hat{y}_i = y_i)$,
where $M$ is the number of individuals for whom the model's \wareki prediction is exactly correct (i.e., $\hat{E}_i = E_i$ and $\hat{x}_i = x_i$).

We also report consistency under \textbf{Within ±3-years Match}, which allows for a 3-year deviation in the Gregorian prediction.
It is defined as:
$\frac{1}{M} \sum_{i=1}^M \mathbbm{1}(|\hat{y}_i - y_i| \leq 3)$.

\begin{figure}[t!]
\centering
\includegraphics[width=0.6\linewidth]{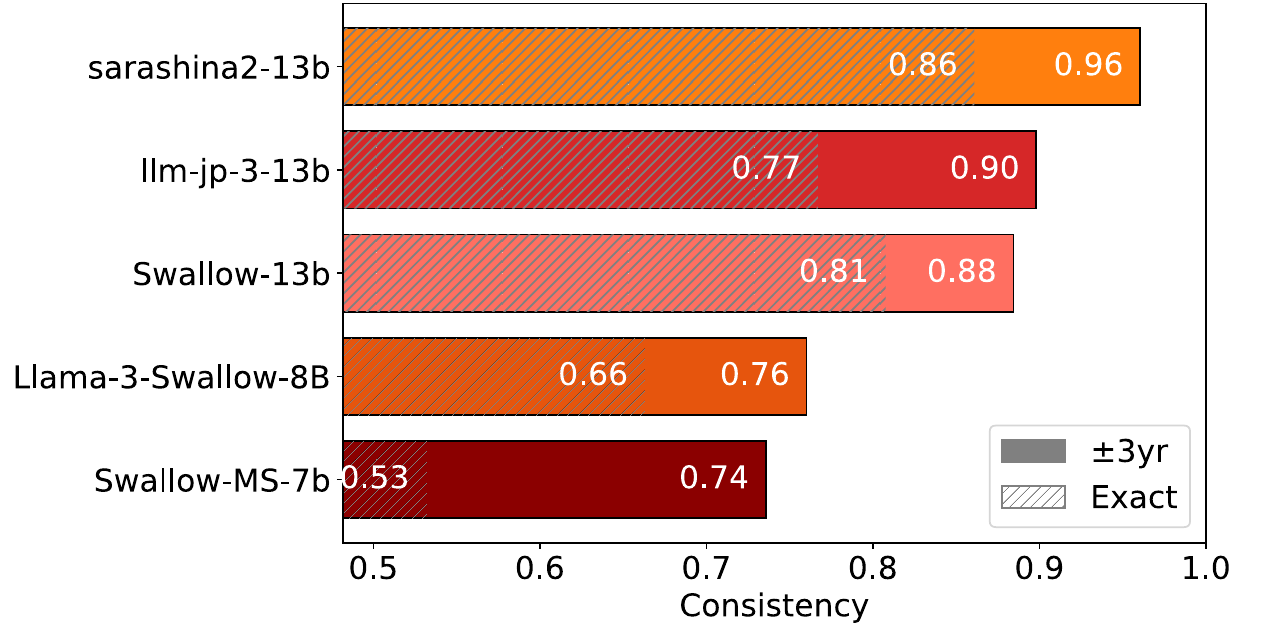}
\caption{
\textsc{CrossCalendarConsistency}: proportion of items correctly answered in the Japanese calendar that are also correctly answered in the Gregorian calendar under exact match and within a $\pm$3-year tolerance.
For exact matches, some Japanese-centric models, such as sarashina2-13b and Swallow-13b, exhibit high consistency above 80\%, while others, like Swallow-MS-7b, remain around 50\%.
}
\label{fig:consistency}
\end{figure}

\paragraph{Results.}
\cref{fig:consistency} shows the consistency of Japanese-centric models in \textsc{BirthYearRecall}, measuring the ratio of cases where the model correctly answers both in \wareki and Gregorian format.
Results are reported using both exact match and 3-year tolerance criteria.
English-centric models are omitted because they rarely produced correct \wareki output in this task.

While some Japanese models, such as sarashina2-13b and Swallow-13b, achieve over 80\% consistency, others like Swallow-MS-7b show lower consistency around 50\%.
These results show that even in Japanese-centric LMs, there is considerable variation in how knowledge relating to Gregorian and Japanese calendars is recalled.

\clearpage

\end{document}

%% file: _default_preamble.tex
\usepackage[preprint]{acl}

\usepackage{times}
\usepackage{latexsym}

\usepackage[T1]{fontenc}

\usepackage[utf8]{inputenc}

\usepackage{microtype}

\usepackage{inconsolata}

\usepackage{graphicx}

%% file: _custom_preamble.tex
\usepackage[whole]{bxcjkjatype}
\usepackage{xspace}

\usepackage[T1]{fontenc}

\definecolor{snsblue}{RGB}{59, 117, 175}
\definecolor{snsgreen}{RGB}{81, 158, 62}
\definecolor{snsorange}{RGB}{239, 134, 53}
\definecolor{snsred}{RGB}{234, 51, 35}
\definecolor{heatmapred}{RGB}{214, 38, 40}
\definecolor{heatmapblue}{HTML}{2077B4}
\definecolor{heatmappurple}{RGB}{140, 70, 160}

\newcommand{\wareki}{\emph{wareki}\xspace}

\usepackage{amsmath}
\usepackage{amssymb}
\usepackage{amsfonts}
\usepackage{bbm}

\usepackage[capitalise,english,nameinlink]{cleveref} %

\crefname{figure}{\text{Fig.}}{\text{Fig.}}
\crefname{section}{\S}{\S\S}
\crefname{equation}{\text{式}}{\text{式}}
\crefname{table}{\text{Tbl.}}{\text{Tbl.}}
\creflabelformat{equation}{#2\textup{#1}#3}
\crefname{appendix}{\mbox{App.}}{\mbox{App.}}

\usepackage{cite}

\usepackage{booktabs}
\usepackage{bm}
\usepackage{tabularx}
\usepackage{arydshln}

%% file: main.bbl
\begin{thebibliography}{23}
\providecommand{\natexlab}[1]{#1}

\bibitem[{Bhatia et~al.(2025{\natexlab{a}})Bhatia, Peyrard, and Zhao}]{bhatia2025datefragmentshiddenbottleneck}
Gagan Bhatia, Maxime Peyrard, and Wei Zhao. 2025{\natexlab{a}}.
\newblock \href {https://arxiv.org/abs/2505.16088} {Date fragments: A hidden bottleneck of tokenization for temporal reasoning}.
\newblock \emph{Preprint}, arXiv:2505.16088.

\bibitem[{Bhatia et~al.(2025{\natexlab{b}})Bhatia, Tang, Mahanta, and Kazi}]{bhatia-etal-2025-datelogicqa}
Gagan Bhatia, Ming~Ze Tang, Cristina Mahanta, and Madiha Kazi. 2025{\natexlab{b}}.
\newblock \href {https://doi.org/10.18653/v1/2025.naacl-srw.32} {{D}ate{L}ogic{QA}: Benchmarking temporal biases in large language models}.
\newblock In \emph{Proceedings of the 2025 Conference of the Nations of the Americas Chapter of the Association for Computational Linguistics: Human Language Technologies (Volume 4: Student Research Workshop)}, pages 321--332, Albuquerque, USA. Association for Computational Linguistics.

\bibitem[{Chen et~al.(2021)Chen, Wang, Wang, and Wang}]{NEURIPS_DATASETS_AND_BENCHMARKS2021_1f0e3dad}
Wenhu Chen, Xinyi Wang, William~Yang Wang, and William~Yang Wang. 2021.
\newblock \href {https://datasets-benchmarks-proceedings.neurips.cc/paper_files/paper/2021/file/1f0e3dad99908345f7439f8ffabdffc4-Paper-round2.pdf} {A dataset for answering time-sensitive questions}.
\newblock In \emph{Proceedings of the Neural Information Processing Systems Track on Datasets and Benchmarks}, volume~1.

\bibitem[{Chu et~al.(2024)Chu, Chen, Chen, Yu, Wang, Liu, and Qin}]{chu-etal-2024-timebench}
Zheng Chu, Jingchang Chen, Qianglong Chen, Weijiang Yu, Haotian Wang, Ming Liu, and Bing Qin. 2024.
\newblock \href {https://doi.org/10.18653/v1/2024.acl-long.66} {{T}ime{B}ench: A comprehensive evaluation of temporal reasoning abilities in large language models}.
\newblock In \emph{Proceedings of the 62nd Annual Meeting of the Association for Computational Linguistics (Volume 1: Long Papers)}, pages 1204--1228, Bangkok, Thailand. Association for Computational Linguistics.

\bibitem[{DeepSeek-AI et~al.(2025)DeepSeek-AI, Liu, Feng, Xue, Wang, Wu, Lu, Zhao, Deng, Zhang, Ruan, Dai, Guo, Yang, Chen, Ji, Li, Lin, Dai, Luo, Hao, Chen, Li, Zhang, Bao, Xu, Wang, Zhang, Ding, Xin, Gao, Li, Qu, Cai, Liang, Guo, Ni, Li, Wang, Chen, Chen, Yuan, Qiu, Li, Song, Dong, Hu, Gao, Guan, Huang, Yu, Wang, Zhang, Xu, Xia, Zhao, Wang, Zhang, Li, Wang, Zhang, Zhang, Tang, Li, Tian, Huang, Wang, Zhang, Wang, Zhu, Chen, Du, Chen, Jin, Ge, Zhang, Pan, Wang, Xu, Zhang, Chen, Li, Lu, Zhou, Chen, Wu, Ye, Ye, Ma, Wang, Zhou, Yu, Zhou, Pan, Wang, Yun, Pei, Sun, Xiao, Zeng, Zhao, An, Liu, Liang, Gao, Yu, Zhang, Li, Jin, Wang, Bi, Liu, Wang, Shen, Chen, Zhang, Chen, Nie, Sun, Wang, Cheng, Liu, Xie, Liu, Yu, Song, Shan, Zhou, Yang, Li, Su, Lin, Li, Wang, Wei, Zhu, Zhang, Xu, Xu, Huang, Li, Zhao, Sun, Li, Wang, Yu, Zheng, Zhang, Shi, Xiong, He, Tang, Piao, Wang, Tan, Ma, Liu, Guo, Wu, Ou, Zhu, Wang, Gong, Zou, He, Zha, Xiong, Ma, Yan, Luo, You, Liu, Zhou, Wu, Ren, Ren, Sha, Fu, Xu, Huang, Zhang, Xie, Zhang, Hao,
  Gou, Ma, Yan, Shao, Xu, Wu, Zhang, Li, Gu, Zhu, Liu, Li, Xie, Song, Gao, and Pan}]{deepseekai2025deepseekv3technicalreport}
DeepSeek-AI, Aixin Liu, Bei Feng, Bing Xue, Bingxuan Wang, Bochao Wu, Chengda Lu, Chenggang Zhao, Chengqi Deng, Chenyu Zhang, Chong Ruan, Damai Dai, Daya Guo, Dejian Yang, Deli Chen, Dongjie Ji, Erhang Li, Fangyun Lin, Fucong Dai, and 181 others. 2025.
\newblock \href {https://arxiv.org/abs/2412.19437} {Deepseek-v3 technical report}.
\newblock \emph{Preprint}, arXiv:2412.19437.

\bibitem[{El-Shangiti et~al.(2025)El-Shangiti, Hiraoka, AlQuabeh, Heinzerling, and Inui}]{el-shangiti-etal-2025-geometry}
Ahmed~Oumar El-Shangiti, Tatsuya Hiraoka, Hilal AlQuabeh, Benjamin Heinzerling, and Kentaro Inui. 2025.
\newblock \href {https://doi.org/10.18653/v1/2025.naacl-short.47} {The geometry of numerical reasoning: Language models compare numeric properties in linear subspaces}.
\newblock In \emph{Proceedings of the 2025 Conference of the Nations of the Americas Chapter of the Association for Computational Linguistics: Human Language Technologies (Volume 2: Short Papers)}, pages 550--561, Albuquerque, New Mexico. Association for Computational Linguistics.

\bibitem[{Enomoto et~al.(2024)Enomoto, Tolmachev, Niitsuma, Kurita, and Kawahara}]{enomoto-etal-2024-investigating}
Rintaro Enomoto, Arseny Tolmachev, Takuro Niitsuma, Shuhei Kurita, and Daisuke Kawahara. 2024.
\newblock \href {https://doi.org/10.18653/v1/2024.naacl-srw.18} {Investigating web corpus filtering methods for language model development in {J}apanese}.
\newblock In \emph{Proceedings of the 2024 Conference of the North American Chapter of the Association for Computational Linguistics: Human Language Technologies (Volume 4: Student Research Workshop)}, pages 154--160, Mexico City, Mexico. Association for Computational Linguistics.

\bibitem[{Fujii et~al.(2024)Fujii, Nakamura, Loem, Iida, Ohi, Hattori, Shota, Mizuki, Yokota, and Okazaki}]{swallow_01}
Kazuki Fujii, Taishi Nakamura, Mengsay Loem, Hiroki Iida, Masanari Ohi, Kakeru Hattori, Hirai Shota, Sakae Mizuki, Rio Yokota, and Naoaki Okazaki. 2024.
\newblock \href {https://openreview.net/forum?id=TQdd1VhWbe} {{Continual Pre-Training for Cross-Lingual {LLM} Adaptation: Enhancing Japanese Language Capabilities}}.
\newblock In \emph{Proceedings of the First Conference on Language Modeling (COLM)}.

\bibitem[{Gaere and Wangenheim(2025)}]{gaere2025datetimenewbenchmarkmeasure}
Edward Gaere and Florian Wangenheim. 2025.
\newblock \href {https://arxiv.org/abs/2504.16155} {Datetime: A new benchmark to measure llm translation and reasoning capabilities}.
\newblock \emph{Preprint}, arXiv:2504.16155.

\bibitem[{Grattafiori et~al.(2024)Grattafiori, Dubey, Jauhri, Pandey, Kadian, Al-Dahle, Letman, Mathur, Schelten, Vaughan, Yang, Fan, Goyal, Hartshorn, Yang, Mitra, Sravankumar, Korenev, Hinsvark, Rao, Zhang, Rodriguez, Gregerson, Spataru, Roziere, Biron, Tang, Chern, Caucheteux, Nayak, Bi, Marra, McConnell, Keller, Touret, Wu, Wong, Ferrer, Nikolaidis, Allonsius, Song, Pintz, Livshits, Wyatt, Esiobu, Choudhary, Mahajan, Garcia-Olano, Perino, Hupkes, Lakomkin, AlBadawy, Lobanova, Dinan, Smith, Radenovic, Guzmán, Zhang, Synnaeve, Lee, Anderson, Thattai, Nail, Mialon, Pang, Cucurell, Nguyen, Korevaar, Xu, Touvron, Zarov, Ibarra, Kloumann, Misra, Evtimov, Zhang, Copet, Lee, Geffert, Vranes, Park, Mahadeokar, Shah, van~der Linde, Billock, Hong, Lee, Fu, Chi, Huang, Liu, Wang, Yu, Bitton, Spisak, Park, Rocca, Johnstun, Saxe, Jia, Alwala, Prasad, Upasani, Plawiak, Li, Heafield, Stone, El-Arini, Iyer, Malik, Chiu, Bhalla, Lakhotia, Rantala-Yeary, van~der Maaten, Chen, Tan, Jenkins, Martin, Madaan, Malo, Blecher,
  Landzaat, de~Oliveira, Muzzi, Pasupuleti, Singh, Paluri, Kardas, Tsimpoukelli, Oldham, Rita, Pavlova, Kambadur, Lewis, Si, Singh, Hassan, Goyal, Torabi, Bashlykov, Bogoychev, Chatterji, Zhang, Duchenne, Çelebi, Alrassy, Zhang, Li, Vasic, Weng, Bhargava, Dubal, Krishnan, Koura, Xu, He, Dong, Srinivasan, Ganapathy, Calderer, Cabral, Stojnic, Raileanu, Maheswari, Girdhar, Patel, Sauvestre, Polidoro, Sumbaly, Taylor, Silva, Hou, Wang, Hosseini, Chennabasappa, Singh, Bell, Kim, Edunov, Nie, Narang, Raparthy, Shen, Wan, Bhosale, Zhang, Vandenhende, Batra, Whitman, Sootla, Collot, Gururangan, Borodinsky, Herman, Fowler, Sheasha, Georgiou, Scialom, Speckbacher, Mihaylov, Xiao, Karn, Goswami, Gupta, Ramanathan, Kerkez, Gonguet, Do, Vogeti, Albiero, Petrovic, Chu, Xiong, Fu, Meers, Martinet, Wang, Wang, Tan, Xia, Xie, Jia, Wang, Goldschlag, Gaur, Babaei, Wen, Song, Zhang, Li, Mao, Coudert, Yan, Chen, Papakipos, Singh, Srivastava, Jain, Kelsey, Shajnfeld, Gangidi, Victoria, Goldstand, Menon, Sharma, Boesenberg,
  Baevski, Feinstein, Kallet, Sangani, Teo, Yunus, Lupu, Alvarado, Caples, Gu, Ho, Poulton, Ryan, Ramchandani, Dong, Franco, Goyal, Saraf, Chowdhury, Gabriel, Bharambe, Eisenman, Yazdan, James, Maurer, Leonhardi, Huang, Loyd, Paola, Paranjape, Liu, Wu, Ni, Hancock, Wasti, Spence, Stojkovic, Gamido, Montalvo, Parker, Burton, Mejia, Liu, Wang, Kim, Zhou, Hu, Chu, Cai, Tindal, Feichtenhofer, Gao, Civin, Beaty, Kreymer, Li, Adkins, Xu, Testuggine, David, Parikh, Liskovich, Foss, Wang, Le, Holland, Dowling, Jamil, Montgomery, Presani, Hahn, Wood, Le, Brinkman, Arcaute, Dunbar, Smothers, Sun, Kreuk, Tian, Kokkinos, Ozgenel, Caggioni, Kanayet, Seide, Florez, Schwarz, Badeer, Swee, Halpern, Herman, Sizov, Guangyi, Zhang, Lakshminarayanan, Inan, Shojanazeri, Zou, Wang, Zha, Habeeb, Rudolph, Suk, Aspegren, Goldman, Zhan, Damlaj, Molybog, Tufanov, Leontiadis, Veliche, Gat, Weissman, Geboski, Kohli, Lam, Asher, Gaya, Marcus, Tang, Chan, Zhen, Reizenstein, Teboul, Zhong, Jin, Yang, Cummings, Carvill, Shepard, McPhie,
  Torres, Ginsburg, Wang, Wu, U, Saxena, Khandelwal, Zand, Matosich, Veeraraghavan, Michelena, Li, Jagadeesh, Huang, Chawla, Huang, Chen, Garg, A, Silva, Bell, Zhang, Guo, Yu, Moshkovich, Wehrstedt, Khabsa, Avalani, Bhatt, Mankus, Hasson, Lennie, Reso, Groshev, Naumov, Lathi, Keneally, Liu, Seltzer, Valko, Restrepo, Patel, Vyatskov, Samvelyan, Clark, Macey, Wang, Hermoso, Metanat, Rastegari, Bansal, Santhanam, Parks, White, Bawa, Singhal, Egebo, Usunier, Mehta, Laptev, Dong, Cheng, Chernoguz, Hart, Salpekar, Kalinli, Kent, Parekh, Saab, Balaji, Rittner, Bontrager, Roux, Dollar, Zvyagina, Ratanchandani, Yuvraj, Liang, Alao, Rodriguez, Ayub, Murthy, Nayani, Mitra, Parthasarathy, Li, Hogan, Battey, Wang, Howes, Rinott, Mehta, Siby, Bondu, Datta, Chugh, Hunt, Dhillon, Sidorov, Pan, Mahajan, Verma, Yamamoto, Ramaswamy, Lindsay, Lindsay, Feng, Lin, Zha, Patil, Shankar, Zhang, Zhang, Wang, Agarwal, Sajuyigbe, Chintala, Max, Chen, Kehoe, Satterfield, Govindaprasad, Gupta, Deng, Cho, Virk, Subramanian, Choudhury,
  Goldman, Remez, Glaser, Best, Koehler, Robinson, Li, Zhang, Matthews, Chou, Shaked, Vontimitta, Ajayi, Montanez, Mohan, Kumar, Mangla, Ionescu, Poenaru, Mihailescu, Ivanov, Li, Wang, Jiang, Bouaziz, Constable, Tang, Wu, Wang, Wu, Gao, Kleinman, Chen, Hu, Jia, Qi, Li, Zhang, Zhang, Adi, Nam, Yu, Wang, Zhao, Hao, Qian, Li, He, Rait, DeVito, Rosnbrick, Wen, Yang, Zhao, and Ma}]{grattafiori2024Llama3herdmodels}
Aaron Grattafiori, Abhimanyu Dubey, Abhinav Jauhri, Abhinav Pandey, Abhishek Kadian, Ahmad Al-Dahle, Aiesha Letman, Akhil Mathur, Alan Schelten, Alex Vaughan, Amy Yang, Angela Fan, Anirudh Goyal, Anthony Hartshorn, Aobo Yang, Archi Mitra, Archie Sravankumar, Artem Korenev, Arthur Hinsvark, and 542 others. 2024.
\newblock \href {https://arxiv.org/abs/2407.21783} {The llama 3 herd of models}.
\newblock \emph{Preprint}, arXiv:2407.21783.

\bibitem[{Heinzerling and Inui(2024)}]{heinzerling-inui-2024-monotonic}
Benjamin Heinzerling and Kentaro Inui. 2024.
\newblock \href {https://doi.org/10.18653/v1/2024.acl-short.18} {Monotonic representation of numeric attributes in language models}.
\newblock In \emph{Proceedings of the 62nd Annual Meeting of the Association for Computational Linguistics (Volume 2: Short Papers)}, pages 175--195, Bangkok, Thailand. Association for Computational Linguistics.

\bibitem[{Jiang et~al.(2023)Jiang, Sablayrolles, Mensch, Bamford, Chaplot, de~las Casas, Bressand, Lengyel, Lample, Saulnier, Lavaud, Lachaux, Stock, Scao, Lavril, Wang, Lacroix, and Sayed}]{jiang2023mistral7b}
Albert~Q. Jiang, Alexandre Sablayrolles, Arthur Mensch, Chris Bamford, Devendra~Singh Chaplot, Diego de~las Casas, Florian Bressand, Gianna Lengyel, Guillaume Lample, Lucile Saulnier, Lélio~Renard Lavaud, Marie-Anne Lachaux, Pierre Stock, Teven~Le Scao, Thibaut Lavril, Thomas Wang, Timothée Lacroix, and William~El Sayed. 2023.
\newblock \href {https://arxiv.org/abs/2310.06825} {Mistral 7b}.
\newblock \emph{Preprint}, arXiv:2310.06825.

\bibitem[{Khairallah et~al.(2024)Khairallah, Khalifa, Marzouk, Nassar, and Habash}]{khairallah-etal-2024-camel}
Christian Khairallah, Salam Khalifa, Reham Marzouk, Mayar Nassar, and Nizar Habash. 2024.
\newblock \href {https://aclanthology.org/2024.lrec-main.240/} {Camel morph {MSA}: A large-scale open-source morphological analyzer for {M}odern {S}tandard {A}rabic}.
\newblock In \emph{Proceedings of the 2024 Joint International Conference on Computational Linguistics, Language Resources and Evaluation (LREC-COLING 2024)}, pages 2683--2691, Torino, Italia. ELRA and ICCL.

\bibitem[{Kim et~al.(2024)Kim, Suk, Oh, Yoo, Thorne, and Oh}]{kim-etal-2024-click}
Eunsu Kim, Juyoung Suk, Philhoon Oh, Haneul Yoo, James Thorne, and Alice Oh. 2024.
\newblock \href {https://aclanthology.org/2024.lrec-main.296/} {{CLI}c{K}: A benchmark dataset of cultural and linguistic intelligence in {K}orean}.
\newblock In \emph{Proceedings of the 2024 Joint International Conference on Computational Linguistics, Language Resources and Evaluation (LREC-COLING 2024)}, pages 3335--3346, Torino, Italia. ELRA and ICCL.

\bibitem[{Liu et~al.(2024)Liu, Min, Zettlemoyer, Choi, and Hajishirzi}]{Liu2024InfiniGram}
Jiacheng Liu, Sewon Min, Luke Zettlemoyer, Yejin Choi, and Hannaneh Hajishirzi. 2024.
\newblock \href {https://openreview.net/forum?id=u2vAyMeLMm} {Infini-gram: Scaling unbounded n-gram language models to a trillion tokens}.
\newblock In \emph{First Conference on Language Modeling}.

\bibitem[{LLM-jp(2024)}]{llmjp2024llmjpcrossorganizationalprojectresearch}
LLM-jp. 2024.
\newblock \href {https://arxiv.org/abs/2407.03963} {Llm-jp: A cross-organizational project for the research and development of fully open japanese llms}.
\newblock \emph{Preprint}, arXiv:2407.03963.

\bibitem[{Okazaki et~al.(2024)Okazaki, Hattori, Shota, Iida, Ohi, Fujii, Nakamura, Loem, Yokota, and Mizuki}]{swallow_02}
Naoaki Okazaki, Kakeru Hattori, Hirai Shota, Hiroki Iida, Masanari Ohi, Kazuki Fujii, Taishi Nakamura, Mengsay Loem, Rio Yokota, and Sakae Mizuki. 2024.
\newblock \href {https://openreview.net/forum?id=N5EYQSwW26} {{Building a Large Japanese Web Corpus for Large Language Models}}.
\newblock In \emph{Proceedings of the First Conference on Language Modeling (COLM)}.

\bibitem[{OpenAI et~al.(2024)OpenAI, :, Hurst, Lerer, Goucher, Perelman, Ramesh, Clark, Ostrow, Welihinda, Hayes, Radford, Mądry, Baker-Whitcomb, Beutel, Borzunov, Carney, Chow, Kirillov, Nichol, Paino, Renzin, Passos, Kirillov, Christakis, Conneau, Kamali, Jabri, Moyer, Tam, Crookes, Tootoochian, Tootoonchian, Kumar, Vallone, Karpathy, Braunstein, Cann, Codispoti, Galu, Kondrich, Tulloch, Mishchenko, Baek, Jiang, Pelisse, Woodford, Gosalia, Dhar, Pantuliano, Nayak, Oliver, Zoph, Ghorbani, Leimberger, Rossen, Sokolowsky, Wang, Zweig, Hoover, Samic, McGrew, Spero, Giertler, Cheng, Lightcap, Walkin, Quinn, Guarraci, Hsu, Kellogg, Eastman, Lugaresi, Wainwright, Bassin, Hudson, Chu, Nelson, Li, Shern, Conger, Barette, Voss, Ding, Lu, Zhang, Beaumont, Hallacy, Koch, Gibson, Kim, Choi, McLeavey, Hesse, Fischer, Winter, Czarnecki, Jarvis, Wei, Koumouzelis, Sherburn, Kappler, Levin, Levy, Carr, Farhi, Mely, Robinson, Sasaki, Jin, Valladares, Tsipras, Li, Nguyen, Findlay, Oiwoh, Wong, Asdar, Proehl, Yang, Antonow,
  Kramer, Peterson, Sigler, Wallace, Brevdo, Mays, Khorasani, Such, Raso, Zhang, von Lohmann, Sulit, Goh, Oden, Salmon, Starace, Brockman, Salman, Bao, Hu, Wong, Wang, Schmidt, Whitney, Jun, Kirchner, de~Oliveira~Pinto, Ren, Chang, Chung, Kivlichan, O'Connell, O'Connell, Osband, Silber, Sohl, Okuyucu, Lan, Kostrikov, Sutskever, Kanitscheider, Gulrajani, Coxon, Menick, Pachocki, Aung, Betker, Crooks, Lennon, Kiros, Leike, Park, Kwon, Phang, Teplitz, Wei, Wolfe, Chen, Harris, Varavva, Lee, Shieh, Lin, Yu, Weng, Tang, Yu, Jang, Candela, Beutler, Landers, Parish, Heidecke, Schulman, Lachman, McKay, Uesato, Ward, Kim, Huizinga, Sitkin, Kraaijeveld, Gross, Kaplan, Snyder, Achiam, Jiao, Lee, Zhuang, Harriman, Fricke, Hayashi, Singhal, Shi, Karthik, Wood, Rimbach, Hsu, Nguyen, Gu-Lemberg, Button, Liu, Howe, Muthukumar, Luther, Ahmad, Kai, Itow, Workman, Pathak, Chen, Jing, Guy, Fedus, Zhou, Mamitsuka, Weng, McCallum, Held, Ouyang, Feuvrier, Zhang, Kondraciuk, Kaiser, Hewitt, Metz, Doshi, Aflak, Simens, Boyd,
  Thompson, Dukhan, Chen, Gray, Hudnall, Zhang, Aljubeh, Litwin, Zeng, Johnson, Shetty, Gupta, Shah, Yatbaz, Yang, Zhong, Glaese, Chen, Janner, Lampe, Petrov, Wu, Wang, Fradin, Pokrass, Castro, de~Castro, Pavlov, Brundage, Wang, Khan, Murati, Bavarian, Lin, Yesildal, Soto, Gimelshein, Cone, Staudacher, Summers, LaFontaine, Chowdhury, Ryder, Stathas, Turley, Tezak, Felix, Kudige, Keskar, Deutsch, Bundick, Puckett, Nachum, Okelola, Boiko, Murk, Jaffe, Watkins, Godement, Campbell-Moore, Chao, McMillan, Belov, Su, Bak, Bakkum, Deng, Dolan, Hoeschele, Welinder, Tillet, Pronin, Tillet, Dhariwal, Yuan, Dias, Lim, Arora, Troll, Lin, Lopes, Puri, Miyara, Leike, Gaubert, Zamani, Wang, Donnelly, Honsby, Smith, Sahai, Ramchandani, Huet, Carmichael, Zellers, Chen, Chen, Nigmatullin, Cheu, Jain, Altman, Schoenholz, Toizer, Miserendino, Agarwal, Culver, Ethersmith, Gray, Grove, Metzger, Hermani, Jain, Zhao, Wu, Jomoto, Wu, Shuaiqi, Xia, Phene, Papay, Narayanan, Coffey, Lee, Hall, Balaji, Broda, Stramer, Xu, Gogineni,
  Christianson, Sanders, Patwardhan, Cunninghman, Degry, Dimson, Raoux, Shadwell, Zheng, Underwood, Markov, Sherbakov, Rubin, Stasi, Kaftan, Heywood, Peterson, Walters, Eloundou, Qi, Moeller, Monaco, Kuo, Fomenko, Chang, Zheng, Zhou, Manassra, Sheu, Zaremba, Patil, Qian, Kim, Cheng, Zhang, He, Zhang, Jin, Dai, and Malkov}]{openai2024gpt4ocard}
OpenAI, :, Aaron Hurst, Adam Lerer, Adam~P. Goucher, Adam Perelman, Aditya Ramesh, Aidan Clark, AJ~Ostrow, Akila Welihinda, Alan Hayes, Alec Radford, Aleksander Mądry, Alex Baker-Whitcomb, Alex Beutel, Alex Borzunov, Alex Carney, Alex Chow, Alex Kirillov, and 401 others. 2024.
\newblock \href {https://arxiv.org/abs/2410.21276} {Gpt-4o system card}.
\newblock \emph{Preprint}, arXiv:2410.21276.

\bibitem[{Pawar et~al.(2024)Pawar, Park, Jin, Arora, Myung, Yadav, Haznitrama, Song, Oh, and Augenstein}]{pawar2024surveyculturalawarenesslanguage}
Siddhesh Pawar, Junyeong Park, Jiho Jin, Arnav Arora, Junho Myung, Srishti Yadav, Faiz~Ghifari Haznitrama, Inhwa Song, Alice Oh, and Isabelle Augenstein. 2024.
\newblock \href {https://arxiv.org/abs/2411.00860} {Survey of cultural awareness in language models: Text and beyond}.
\newblock \emph{Preprint}, arXiv:2411.00860.

\bibitem[{Shen et~al.(2024)Shen, Logeswaran, Lee, Lee, Poria, and Mihalcea}]{shen-etal-2024-understanding}
Siqi Shen, Lajanugen Logeswaran, Moontae Lee, Honglak Lee, Soujanya Poria, and Rada Mihalcea. 2024.
\newblock \href {https://doi.org/10.18653/v1/2024.naacl-long.316} {Understanding the capabilities and limitations of large language models for cultural commonsense}.
\newblock In \emph{Proceedings of the 2024 Conference of the North American Chapter of the Association for Computational Linguistics: Human Language Technologies (Volume 1: Long Papers)}, pages 5668--5680, Mexico City, Mexico. Association for Computational Linguistics.

\bibitem[{Touvron et~al.(2023)Touvron, Martin, Stone, Albert, Almahairi, Babaei, Bashlykov, Batra, Bhargava, Bhosale, Bikel, Blecher, Ferrer, Chen, Cucurull, Esiobu, Fernandes, Fu, Fu, Fuller, Gao, Goswami, Goyal, Hartshorn, Hosseini, Hou, Inan, Kardas, Kerkez, Khabsa, Kloumann, Korenev, Koura, Lachaux, Lavril, Lee, Liskovich, Lu, Mao, Martinet, Mihaylov, Mishra, Molybog, Nie, Poulton, Reizenstein, Rungta, Saladi, Schelten, Silva, Smith, Subramanian, Tan, Tang, Taylor, Williams, Kuan, Xu, Yan, Zarov, Zhang, Fan, Kambadur, Narang, Rodriguez, Stojnic, Edunov, and Scialom}]{touvron2023llama2openfoundation}
Hugo Touvron, Louis Martin, Kevin Stone, Peter Albert, Amjad Almahairi, Yasmine Babaei, Nikolay Bashlykov, Soumya Batra, Prajjwal Bhargava, Shruti Bhosale, Dan Bikel, Lukas Blecher, Cristian~Canton Ferrer, Moya Chen, Guillem Cucurull, David Esiobu, Jude Fernandes, Jeremy Fu, Wenyin Fu, and 49 others. 2023.
\newblock \href {https://arxiv.org/abs/2307.09288} {Llama 2: Open foundation and fine-tuned chat models}.
\newblock \emph{Preprint}, arXiv:2307.09288.

\bibitem[{Wang et~al.(2024)Wang, Yeo, Lim, and Kim}]{wang-etal-2024-kulture}
Xiaonan Wang, Jinyoung Yeo, Joon-Ho Lim, and Hansaem Kim. 2024.
\newblock \href {https://aclanthology.org/2024.paclic-1.88/} {{KULTURE} bench: A benchmark for assessing language model in {K}orean cultural context}.
\newblock In \emph{Proceedings of the 38th Pacific Asia Conference on Language, Information and Computation}, pages 914--927, Tokyo, Japan. Tokyo University of Foreign Studies.

\bibitem[{Wang and Zhao(2024)}]{wang-zhao-2024-tram}
Yuqing Wang and Yun Zhao. 2024.
\newblock \href {https://doi.org/10.18653/v1/2024.findings-acl.382} {{TRAM}: Benchmarking temporal reasoning for large language models}.
\newblock In \emph{Findings of the Association for Computational Linguistics: ACL 2024}, pages 6389--6415, Bangkok, Thailand. Association for Computational Linguistics.

\end{thebibliography}
